\documentclass[runningheads]{llncs}
\usepackage{amsmath,amssymb} 

\usepackage{graphicx}
\usepackage{subcaption}
\usepackage{float}
\usepackage{caption}	
\usepackage{lscape}                                         

\usepackage[lined,ruled,linesnumbered]{algorithm2e}

\usepackage{booktabs}                   
\usepackage{multirow}

\usepackage{paralist}
\usepackage{enumitem}

\usepackage{bm}                          
\usepackage{epsfig}                      
\usepackage{graphicx}                  
\usepackage{mathtools}

\usepackage{color}

\usepackage{comment}

\usepackage{url}  
\usepackage[pagebackref=false,breaklinks=true,colorlinks,urlcolor=blue,citecolor=blue,linkcolor=blue,bookmarks=false]{hyperref}
\usepackage[nocompress]{cite}

\usepackage{listings}

\usepackage{xspace}
\usepackage[table]{xcolor}
\usepackage{setspace}

\usepackage{nicefrac}
\usepackage{microtype}
\usepackage[utf8]{inputenc} 
\usepackage[T1]{fontenc}    

\def\etal{et~al\mbox{.}\ }

\newlength\paramargin
\newlength\figmargin
\newlength\tablemargin
\newlength\secmargin
\newlength\figcapmargin
\newlength\rowmargin

\newlength\twoimg

\newcommand{\red}{\textcolor{red}}
\newcommand{\blue}{\textcolor{blue}}

\definecolor{napiergreen}{rgb}{0.16, 0.5, 0.0}
\newcommand{\green}{\textcolor{napiergreen}}

\newcommand{\mpage}[2]
{
\begin{minipage}{#1\linewidth}\centering
#2
\end{minipage}
}

\newcommand{\figref}[1]{Figure~\ref{fig:#1}}

\long\def\ignorethis#1{}

\newcommand {\esther}[1]{}

\newcommand{\first}[1]{{\color{blue}\textbf{#1}}}
\newcommand{\second}[1]{{\color{red}\underline{#1}}}

\newcommand{\tb}[1]{\textbf{#1}}



\graphicspath{{figure}, {example}}

\usepackage[width=122mm,left=12mm,paperwidth=146mm,height=193mm,top=12mm,paperheight=217mm]{geometry}

\begin{document}
\pagestyle{headings}
\mainmatter
\def\ECCVSubNumber{3006}  

\title{NAS-DIP: Learning Deep Image Prior with \\ Neural Architecture Search} 

\titlerunning{NAS-DIP: Learning Deep Image Prior with Neural Architecture Search}

\author{
Yun-Chun Chen\thanks{ equal contribution} \and
Chen Gao$^\star$ \and
Esther Robb \and
Jia-Bin Huang
}

\authorrunning{Y.-C. Chen, C. Gao, E. Robb, and J.-B. Huang}

\institute{Virginia Tech}


\maketitle

\begin{center}
  \mpage{0.1795}{\includegraphics[width=\linewidth]{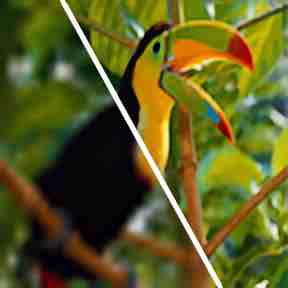}} \hfill
  \mpage{0.1795}{\includegraphics[width=\linewidth]{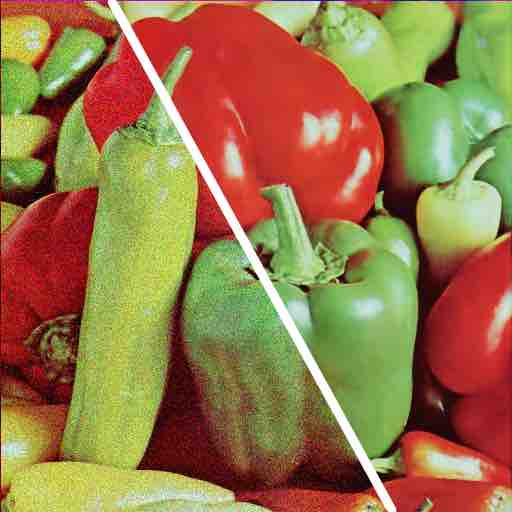}} \hfill
  \mpage{0.1795}{\includegraphics[width=\linewidth]{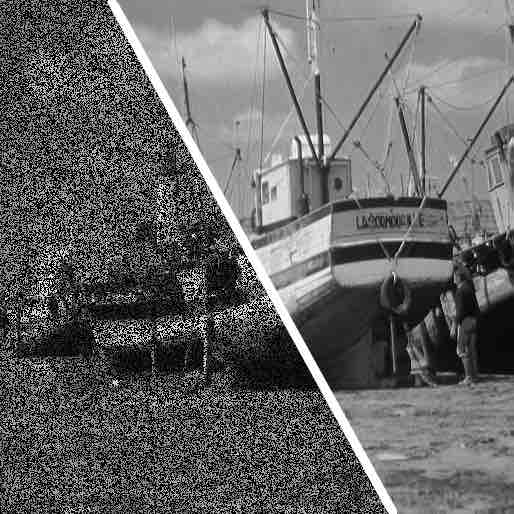}} \hfill
  \mpage{0.1795}{\includegraphics[width=\linewidth]{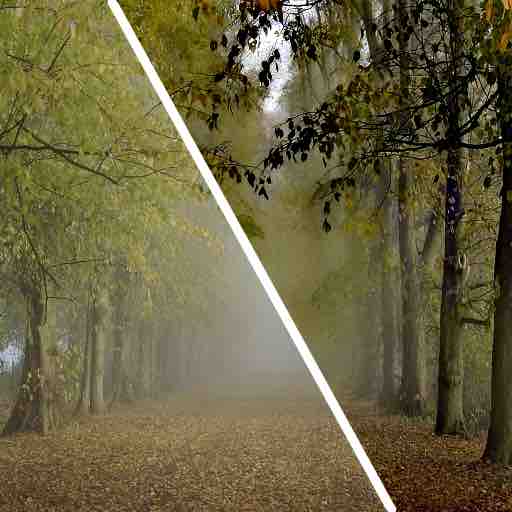}} \hfill
  \mpage{0.1795}{\includegraphics[width=\linewidth]{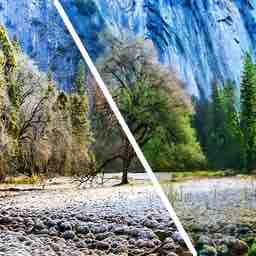}} \\
  \vspace{1.5mm}
  \mpage{0.1795}{Super-Res} \hfill
  \mpage{0.1795}{Denoising} \hfill
  \mpage{0.1795}{Inpainting} \hfill
  \mpage{0.1795}{Dehazing} \hfill
  \mpage{0.1795}{Translation} \\
  \vspace{\figcapmargin}
  \captionof{figure}{
  \textbf{Applications.} 
  We propose to \emph{learn} deep image prior using a neural architecture search.
  The resulting network can be applied to solve various inverse image problems \emph{without} training the model with a large-scale dataset with ground truth.
  Through extensive experimental evaluations, we show that our model compares favorably against existing hand-crafted CNN models for learning-free image restoration tasks and in some cases even reaches competitive performance when compared with recent learning-based models.
  }
  \label{fig:teaser}
  \vspace{0.5\figmargin}
\end{center}

\begin{abstract}
Recent work has shown that the structure of deep convolutional neural networks can be used as a structured image prior for solving various inverse image restoration tasks.
Instead of using hand-designed architectures, we propose to search for neural architectures that capture stronger image priors.
Building upon a generic U-Net architecture, our core contribution lies in designing new search spaces for (1) an upsampling cell and (2) a pattern of cross-scale residual connections.
We search for an improved network by leveraging an existing neural architecture search algorithm (using reinforcement learning with a recurrent neural network controller).
We validate the effectiveness of our method via a wide variety of applications, including image restoration, dehazing, image-to-image translation, and matrix factorization.
Extensive experimental results show that our algorithm performs favorably against state-of-the-art learning-free approaches and reaches competitive performance with existing learning-based methods in some cases.
\end{abstract}

\section{Introduction}

Convolutional neural networks (CNNs) have been successfully applied to many computer vision tasks.
Apart from image recognition tasks, CNNs have also demonstrated strong performance in image restoration and synthesis problems.
The reason behind these successful stories is often attributed to the ability of CNNs to \emph{learn priors} from large-scale datasets (i.e., the priors are embedded in the \emph{parameters}/\emph{weights} of the trained network).
In contrast to existing supervised learning algorithms that require learning the network parameters from labeled datasets, recent studies have discovered that the \emph{structure} of the network by itself is sufficient to capture rich low-level image statistics~\cite{DIP-CVPR-2018,saxe2011random}.
Such structured image priors encoded in the network architecture are critical for image restoration (e.g., single image super-resolution~\cite{lai2017deep,ledig2017photo,lai2018fast} and image denoising~\cite{burger2012image,lefkimmiatis2017non,xie2012image}) and image synthesis (e.g., image-to-image translation~\cite{zhu2017unpaired,isola2017image,huang2018multimodal,lee2018diverse,lee2020drit++}) tasks.

While learning-free methods~\cite{DIP-CVPR-2018} have demonstrated competitive performance on image restoration tasks when compared with learning-based approaches~\cite{ledig2017photo,lai2017deep}, only conventional network architectures such as ResNet~\cite{he2016deep} or U-Net~\cite{ronneberger2015u} have been evaluated. 
There are two important aspects of network designs for these image restoration problems. 
First, while the design of an encoder has been extensively studied~\cite{krizhevsky2012imagenet,simonyan2014very,huang2017densely,he2016deep}, the design of a decoder~\cite{Wojna-2019,Wojna2017TheDI} (the upsampling cell in particular) receives considerably less attention.
Second, as the spatial resolution of the features is progressively reduced along the path of the feature encoder, it is crucial for the network to recover feature maps with higher spatial resolution.
U-Net~\cite{ronneberger2015u} is one popular design to address this issue by concatenating the encoded features at the corresponding encoder layer with the features in the decoder when performing a sequence of up-convolutions. 
Such skip connection patterns, however, are manually designed and fixed for each task at hand.

{\flushleft {\bf Our work.}}
In this paper, we propose to \emph{search} for both (1) the upsampling cells in the decoder and (2) the skip connection patterns between the encoder and the decoder (i.e., cross-level feature connections).
To achieve this, we develop new search spaces for the two components.
First, to search for the upsampling cell, we decompose a typical upsampling operation into two steps:
i) ways of changing the spatial resolution (e.g., bilinear, bicubic~\cite{dong2015image}, or nearest neighbor upsampling) and 
ii) ways of feature transformation (e.g., 2D convolution or 2D transposed convolution~\cite{zeiler2011adaptive,long2015fully,dosovitskiy2015flownet}).
Second, to search for the cross-level connection patterns, we propose to search connection patterns shared across different feature levels in an encoder-decoder network.

Motivated by the Neural Architecture Search (NAS) algorithm~\cite{Zoph-ICLR-2017,Ghiasi-2019,Gong-2019} which has been shown effective in discovering networks with top performance in a large search space, we leverage reinforcement learning (RL) with a recurrent neural network (RNN) controller~\cite{Zoph-ICLR-2017,Ghiasi-2019,Gong-2019} and use the PSNR as the reward to guide the architecture search.
By simultaneously searching in the two developed search spaces, our method is capable of discovering a CNN architecture that captures stronger structured image priors for the task of interest.
We show the applicability of our method through four \emph{learning-free} image restoration tasks, including single image super-resolution, image denoising, image inpainting, and image dehazing, and a \emph{learning-based} unpaired image-to-image translation problem (see Figure~\ref{fig:teaser}).
Our experimental results demonstrate that searching for both the upsampling cell and the cross-level feature connections results in performance improvement over conventional neural architectures.

{\flushleft {\bf Our contributions.}}
First, we present a decomposition based on several commonly used upsampling operators that allows us to search for a novel upsampling cell for each task.
Second, we develop a search space that consists of patterns of cross-level feature connections in an encoder-decoder architecture.
Third, extensive evaluations on a variety of image restoration and synthesis tasks demonstrate that our proposed algorithm compares favorably against existing learning-based methods in some cases and achieves the state-of-the-art performance when compared with existing learning-free approaches. 
\vspace{1.0\secmargin}
\section{Related Work}
\vspace{0.5\secmargin}

{\flushleft {\bf Upsampling cell.}}
The design of the upsampling cell can be categorized into two groups: 1) non-learnable parameter based methods and 2) learnable parameter based approaches.
\emph{Non-learnable} parameter based methods use interpolation to resize the feature maps from lower spatial resolutions to higher spatial resolutions.
Example operators include bilinear/bicubic interpolation~\cite{dong2015image,odena2016deconvolution}, nearest neighbor upsampling~\cite{odena2016deconvolution,jia2017super,berthelot2017began}, and depth-to-space upsampling~\cite{ledig2017photo}.
\emph{Learnable} parameter based approaches \emph{learn} the mappings between feature maps of lower spatial resolutions and higher ones.
Among the design choices, 2D transposed convolution is one of the popular choices for various dense prediction tasks, including semantic segmentation~\cite{long2015fully}, optical flow estimation~\cite{dosovitskiy2015flownet}, depth prediction~\cite{chen2019crdoco}, and image restoration~\cite{lefkimmiatis2017non,sun2018natural} and synthesis~\cite{odena2016deconvolution} problems.
Recent advances include bilinear additive upsampling~\cite{Wojna2017TheDI,Wojna-2019} and CARAFE~\cite{wang2019carafe}.

In contrast to these methods that manually design the upsampling operations, we develop a search space for the upsampling cell by decoupling several existing upsampling operations into methods of changing the spatial resolution and methods of feature transformation, and adopt a Neural Architecture Search algorithm~\cite{Zoph-ICLR-2017,Ghiasi-2019,Gong-2019} (i.e., reinforcement learning with an RNN controller) to \emph{automatically} discover the optimal upsampling cell for each individual task.
We further demonstrate that the discovered upsampling cells can be transferred across tasks with favorable performance compared with the base network architectures.

{\flushleft {\bf NAS applications and search space.}}
NAS algorithms have been successfully applied to various tasks, including image classification~\cite{Real-ICML-2017,Zoph-ICLR-2017,Zoph-ICLR-2018,Real-ICML-2018,pham2018efficient,Liu-ICLR-2019}, semantic segmentation~\cite{Nekrasov-2018,Liu-CVPR-2019,Chen-2018}, object detection~\cite{Ghiasi-2019}, image restoration~\cite{Suganama-2018,Chu-ECCV-2019}, and image generation~\cite{Gong-2019}.
In the context of object detection, NAS-FPN~\cite{Ghiasi-2019} develops a search space that allows the model to learn pyramidal representations by merging cross-scale features for improved detection of multiple objects with different scales and locations.
In the context of semantic segmentation, several methods focus on searching for the encoder architecture~\cite{Liu-CVPR-2019}, the Atrous Spatial Pyramid Pooling module~\cite{Chen-2018}, or the decoder cell for a compact architecture~\cite{Nekrasov-2018}.
In image restoration tasks, existing algorithms aim at discovering a better encoder structure for single image super-resolution~\cite{Chu-ECCV-2019} or asymmetric encoder-decoder architecture for image inpainting and image denoising~\cite{Suganama-2018}.
In image generation problems, AutoGAN~\cite{Gong-2019} adopts a progressive scheme to search for a better generative adversarial network (GAN) architecture~\cite{goodfellow2014generative}.
Other methods searching for the decoder architecture focus on searching for more compact architectures~\cite{Nekrasov-2018} or optimizing cell structures with hand-designed upsampling cells and feature connections~\cite{Suganama-2018,Gong-2019}.

Our focus differs from these methods in two aspects.
First, we develop a search space for the \emph{upsampling cell}, allowing us to discover an optimal upsampling choice for each task at hand.
Second, we search for a pattern of \emph{cross-level feature connections} and share it across different feature levels in an encoder-decoder architecture.
Our cross-level feature connections are different from those in NAS-FPN~\cite{Ghiasi-2019} in that we aim at recovering feature maps with higher spatial resolution in the decoder, whereas NAS-FPN~\cite{Ghiasi-2019} aims at learning pyramidal feature representations for object detection.
By simultaneously searching for both the upsampling cell and the cross-level feature connections, our searched architecture achieves the state-of-the-art performance when compared with existing learning-free approaches on a variety of image restoration tasks, and also reaches competitive results when compared with existing learning-based algorithms.

{\flushleft {\bf NAS algorithms.}}
NAS methods can be grouped into several categories depending on the search algorithm. 
The primary methods are evolution~\cite{angeline1994evolutionary,stanley2009hypercube,stanley2002evolving,Real-ICML-2018,Real-ICML-2017,Liu-ICLR-2018,xie2017genetic,miikkulainen2019evolving}, reinforcement learning~\cite{cai2018efficient,tan2018mnasnet,Zoph-ICLR-2017,Zoph-ICLR-2018,baker2016designing,zhong2018practical}, and differentiable search~\cite{Liu-ICLR-2019,ahmed2018maskconnect}.
Evolutionary search leverages evolutionary algorithms to discover network structures by randomly mutating high-performing candidates from a population of architectures. 
RL-based approaches adopt either Q-learning~\cite{baker2016designing,zhong2018practical} or policy gradient~\cite{cai2018efficient,tan2018mnasnet,Zoph-ICLR-2017,Zoph-ICLR-2018} strategies to train a recurrent network which outputs a sequence of symbols describing a network architecture~\cite{Zoph-ICLR-2017} or a repeatable cell structure~\cite{Zoph-ICLR-2018}.
Differentiable search methods develop a continuous search space and optimize the network architecture using gradient descent, providing time efficiency at the expense of being memory intensive.

In this work, we follow the RL-based approaches and adopt an RL-based search algorithm with an RNN controller \cite{Zoph-ICLR-2017,Gong-2019} to search for the upsampling cell and the cross-level connection patterns.
We note that the search algorithm is not limited to RL-based approaches.
Other alternatives such as evolutionary based methods or differentiable architecture search algorithms can also be applied. 
The focus of our paper lies in the design of the two search spaces.
We leave the development of the search algorithm as future work.
\vspace{1.0\secmargin}
\section{Proposed Method}
\vspace{0.5\secmargin}

In this section, we first provide an overview of the proposed method.
We then describe the two developed search spaces for the upsampling cell and the cross-scale residual connections, respectively.

\setlength{\twoimg}{0.495\textwidth}
\begin{figure}[!t]
  \centering
  \begin{subfigure}[b]{\twoimg}
    \begin{center}
      \includegraphics[width=\linewidth]{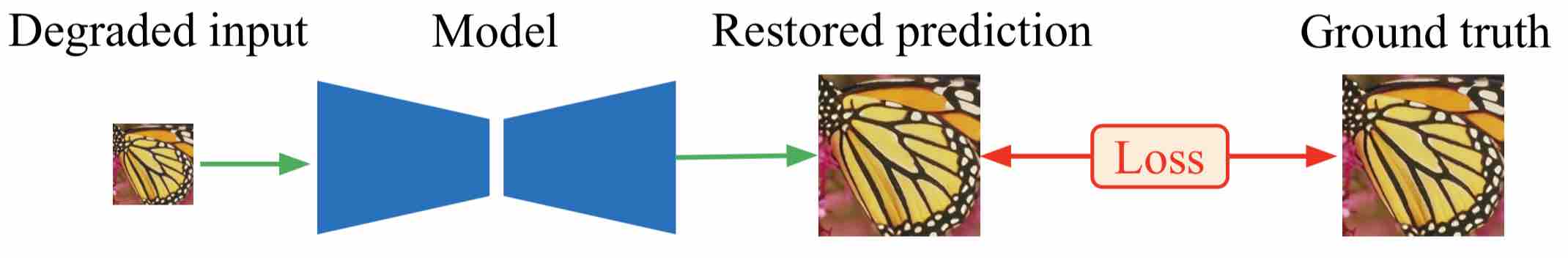}\\
      \caption{Learning-based methods}
      \label{fig:overview-lapsrn}
    \end{center}
    \vfill
    \begin{center}
      \includegraphics[width=\linewidth]{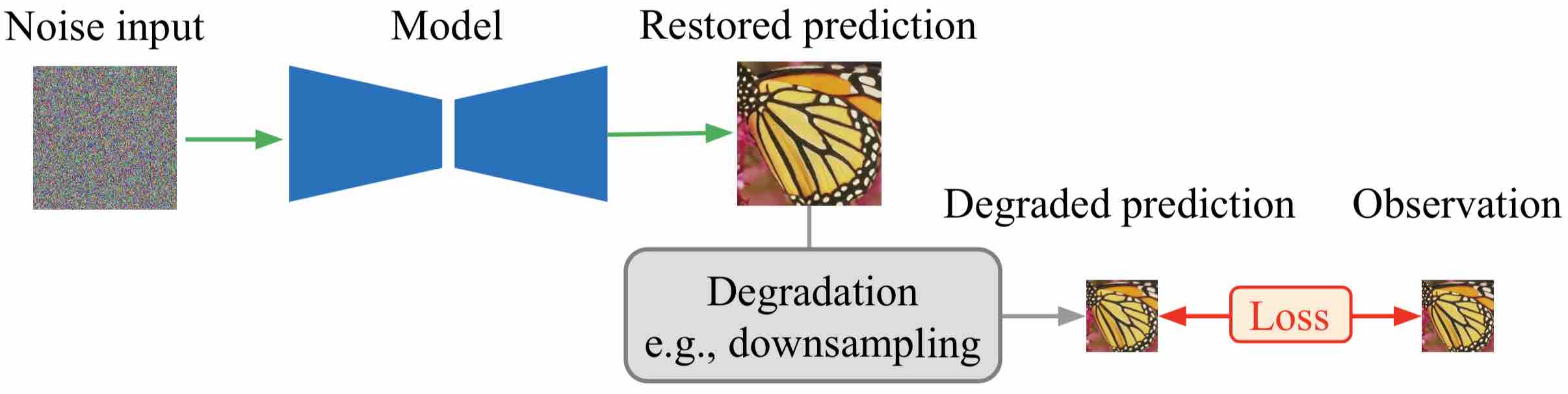}\\
      \caption{Deep Image Prior (DIP)~\cite{DIP-CVPR-2018}}
      \label{fig:overview-dip}
    \end{center}
  \end{subfigure}
  \hfill
  \begin{subfigure}[b]{\twoimg}
    \centering\includegraphics[width=\linewidth]{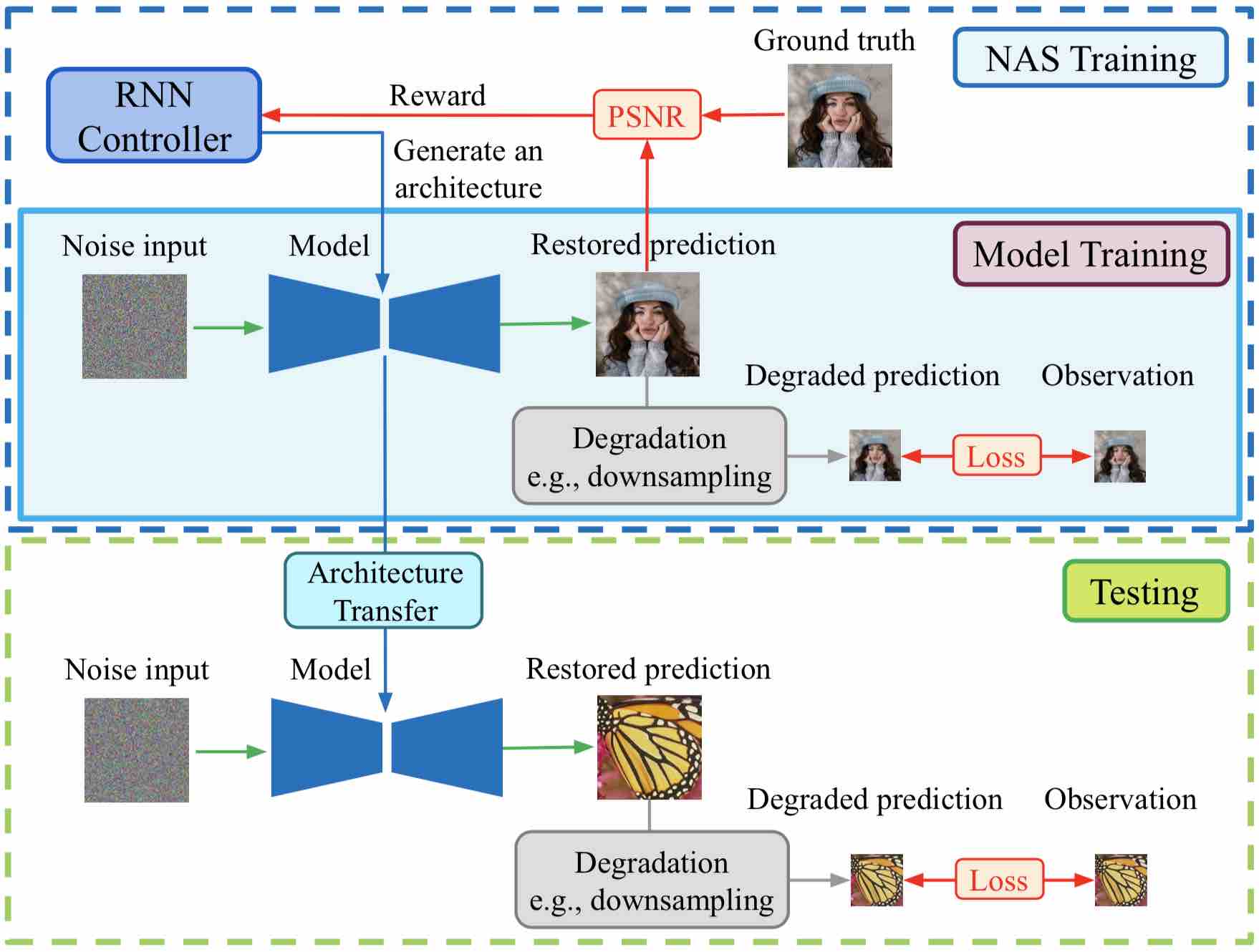}
    \centering\caption{NAS-DIP (Ours)}
    \label{fig:overview-ours}
  \end{subfigure}
  \vspace{2.0\figcapmargin}
  \caption{
  \textbf{Overview of the main workflow and comparison.}
  (a) \tb{Learning-based methods}, e.g., super-resolution models such as LapSRN~\cite{lai2017deep} or VDSR~\cite{kim2016accurate}.
  Given a dataset with labeled input/output pairs (e.g., low-resolution observation and the corresponding high-resolution image), this class of methods trains a deep CNN model to learn the mapping between the degraded input and its corresponding ground truth for the task of interest (e.g., super-resolution, denoising, etc).
  (b) \tb{Learning-free methods.}
  Given a noise input and an input observation, the DIP method~\cite{DIP-CVPR-2018} optimizes the model to produce restored image that matches the given input observation after the specified image degradation model (e.g., downsampling). Here, the weights of the CNN model are \emph{randomly initialized} (i.e., no need to train the model on a labeled dataset).
  (c) \tb{NAS-DIP (Ours).}
  As DIP leverages CNN architectures as structured image priors, we explore ways to \emph{learn} such priors.
  Specifically, we develop two search spaces (one for the upsampling cell and the other for the cross-level feature connections) and leverage existing neural architecture search techniques (an RNN-based controller trained with reinforcement learning) with PSNR as our reward to search for an improved network structure on a held-out training set (blue block).
  After the network architecture search, we then transfer the best-performing architecture and optimize the model the same way as DIP (green block).
  }
  \label{fig:overview}
  \vspace{1.0\figmargin}
\end{figure}

\vspace{\secmargin}
\subsection{Method overview}

In contrast to existing learning-based methods~\cite{lai2017deep} that learn directly from large-scale datasets (\figref{overview-lapsrn}), recent studies~\cite{DIP-CVPR-2018} have shown that by randomly mapping noise to a degenerated (e.g., noisy, low-resolution, or occluded) image, the \emph{untrained} CNN can solve the image restoration problems with competitive performance (\figref{overview-dip}).
To discover network structures that capture stronger image priors, we consider the task of searching for an upsampling cell and a pattern of cross-scale residual connections \emph{without} learning from paired data.
To achieve this, we present an algorithm consisting of two steps.
As shown in \figref{overview-ours}, we first apply reinforcement learning with an RNN controller~\cite{Zoph-ICLR-2017} (using the PSNR as the reward) to search for the best-performing network structure $f_\theta^{*}$ on a held-out training set (blue block).
After the network architecture search step, for each image in the test set, we randomly reinitialize the weights of the best-performing network structure $f_\theta^{*}$ and optimize the mapping from the random noise to a degenerated image (green block).

{\flushleft {\bf Searching for the best-performing network structure.}}
Given an image $x \in \mathbb{R}^{H \times W \times 3}$ in the training set, we first generate a \emph{degenerated} version $x_0$ by adding noise, downsampling, or dropping certain pixels from $x$ depending on the task of interest.
That is, for image denoising, $x_0 \in \mathbb{R}^{H \times W \times 3}$ denotes the noisy version of $x$, for single image super-resolution, $x_0 \in \mathbb{R}^{\frac{H}{r} \times \frac{W}{r} \times 3}$ denotes the low-resolution version of $x$ where $r$ represents the downsampling ratio, and for image inpainting, $x_0 \in \mathbb{R}^{H \times W \times 3}$ denotes the occluded version of $x$.
We then sample a noise image $z \in \mathbb{R}^{H \times W \times C}$ and enforce the searched network $f_\theta$ to map the noise image $z$ to the denoised, high-resolution, or inpainted version of $x_0$, i.e., map the noise image $z$ to $x$.
To achieve this, we follow DIP~\cite{DIP-CVPR-2018} and optimize different objectives for different tasks.

\begin{figure}[!t]
  \begin{center}
  \includegraphics[width=\linewidth]{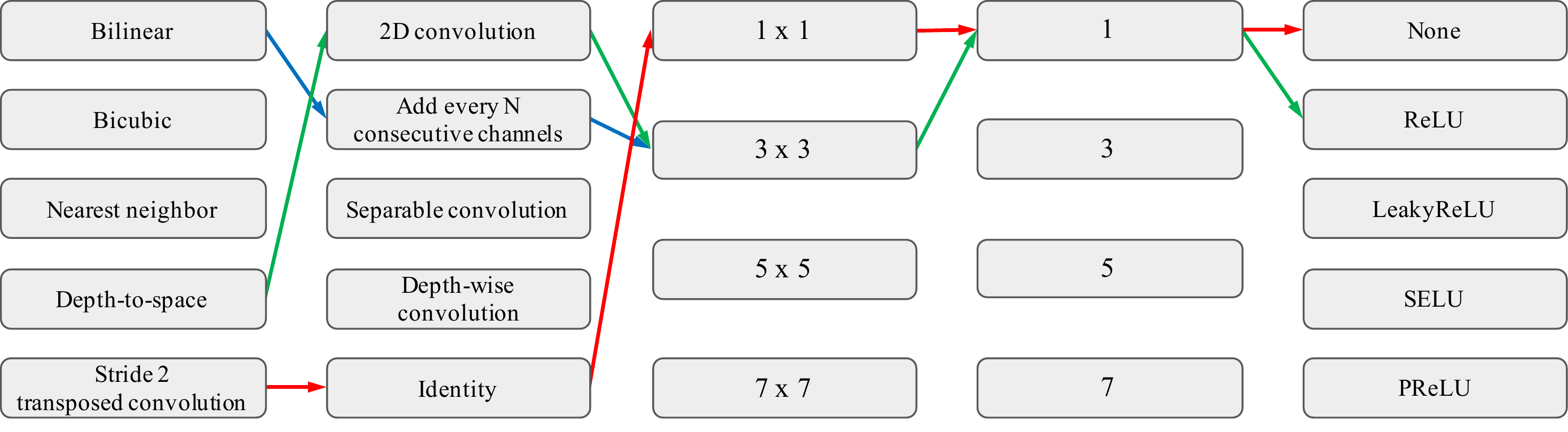} \\
  \mpage{0.159}{\small Upsampling} \hfill
  \mpage{0.0001}{} \hfill
  \mpage{0.159}{\small Transformation} \hfill
  \mpage{0.0001}{} \hfill
  \mpage{0.159}{\small Kernel size} \hfill
  \mpage{0.0001}{} \hfill
  \mpage{0.159}{\small Dilation rate} \hfill
  \mpage{0.0001}{} \hfill
  \mpage{0.159}{\small Activation} \\
  \vspace{0.5\figcapmargin}
  \caption{
  \textbf{Search space for the upsampling cell.} 
  Our search space consists of five main steps (i.e., spatial feature upsampling, feature transformation, kernel size, dilation rate, and activation layer).
  Each step has a set of discrete options.
  Our search space is expressive and covers many existing designs of upsampling operations.
  For example, the \blue{blue} path indicates the bilinear additive upsampling operation~\cite{Wojna2017TheDI,Wojna-2019}.
  The \red{red} path corresponds to the stride $2$ transposed convolution used in~\cite{zeiler2011adaptive,long2015fully}.
  The \green{green} path represents the sub-pixel convolution~\cite{shi2016real,ledig2017photo}. 
  Searching in this space allows us to discover new and effective upsampling layers for capturing stronger image priors.
  }
  \label{fig:upsample}
  \end{center}
  \vspace{1.0\figmargin}
\end{figure}

As the ground-truth images in the training set are available, we can rank each of the searched network structure by computing the Peak Signal to Noise Ratio (PSNR) between the ground-truth image and the network's output (i.e., $f_\theta(z)$) and determine the best-performing network structure $f_\theta^{*}$ for the training set.

{\flushleft {\bf Determining the optimal stopping point $t^*$.}}
We note that the optimal stopping point $t^*$ (i.e., the number of iterations required) depends on the network structure. 
Since we have a held-out training set, we are able to estimate the best stopping point for each randomly generated network structure. 
We then rank all the sampled network structures by measuring the differences (i.e., computing the PSNR) between the recovered image and its corresponding ground truth (i.e., the original image from the training set). 
After that, we apply the best-performing network structure to the test set and report the results recorded at the optimal stopping point $t^*$.

{\flushleft {\bf Testing with the searched network structure $f_\theta^{*}$.}}
After searching on the training set, we apply the best-performing network structure $f_\theta^{*}$ with \emph{random initialization} on each image in the test set at a time for optimization with $t^*$ iterations using the same objective as DIP~\cite{DIP-CVPR-2018} for different tasks.

\begin{figure}[!t]
  \begin{center}
  \mpage{0.3}{\includegraphics[width=\linewidth]{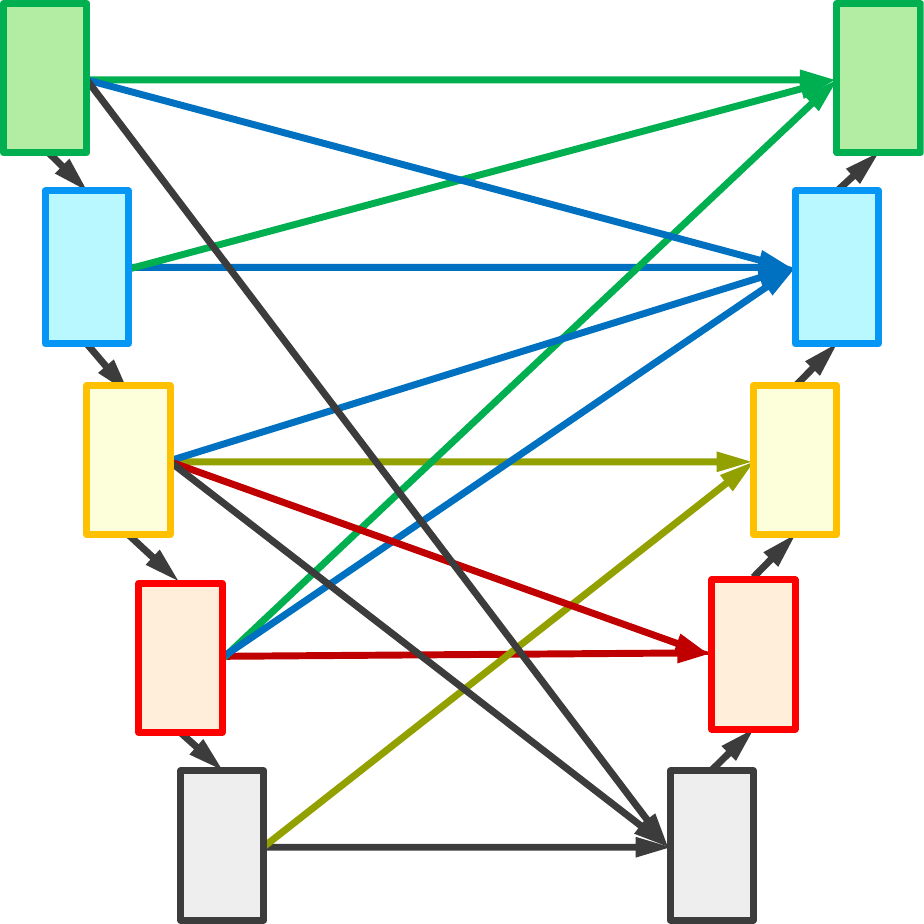}} \hfill
  \mpage{0.3}{\includegraphics[width=\linewidth]{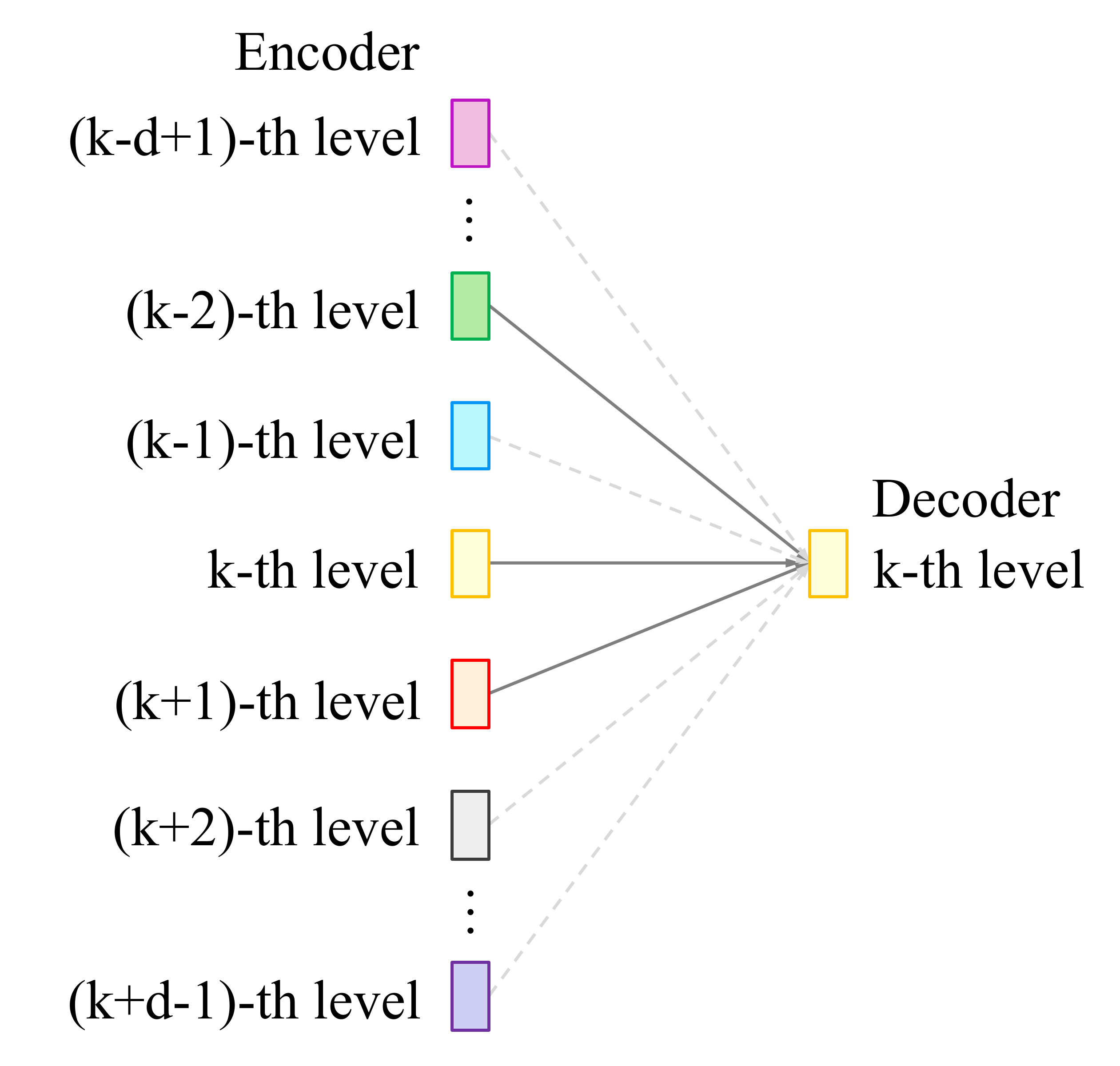}} \hfill
  \mpage{0.3}{\includegraphics[width=\linewidth]{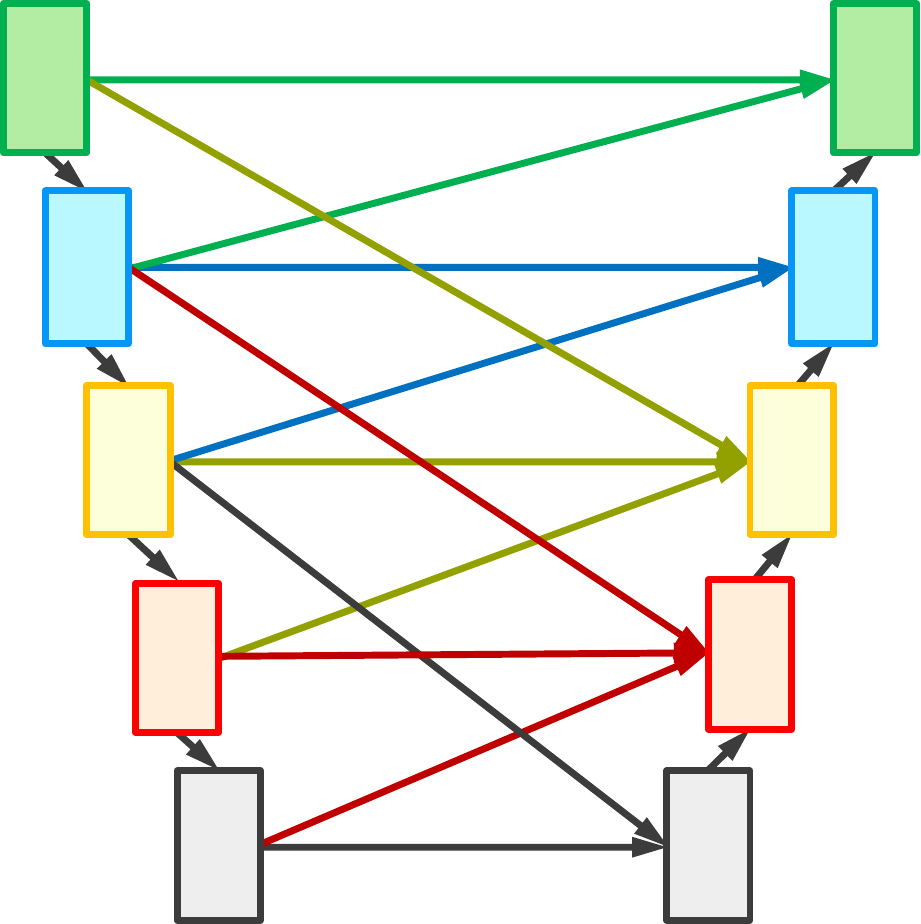}} \\
  \vspace{1.5mm}
  \mpage{0.3}{\small Random pattern} \hfill
  \mpage{0.3}{\small Search space (Ours)} \hfill
  \mpage{0.3}{\small Shared pattern (Ours)} \\
  \vspace{\figcapmargin}
  \caption{
  \textbf{Illustration of the cross-level feature connections.} 
  (\emph{Left}) U-Net architecture with a random pattern of cross-level feature connections. 
  Without any constraint, the search space is large.
  (\emph{Middle}) Our proposed search space for the cross-level feature connections. 
  To constrain the search space, we enforce the connection pattern to depend only on the level difference and share the pattern across different feature levels, e.g., each feature level in the decoder receives feature maps from two levels lower, the same level, and one level higher.
  With this constraint, the size of the search space is significantly reduced.
  (\emph{Right}) U-Net architecture with the pattern of cross-level feature connections shown in the middle example shared across different feature levels.
  }
  \label{fig:connection-illustration}
  \end{center}
  \vspace{\figmargin}
\end{figure}

\subsection{Search space for the upsampling layer}

We develop a search space for the upsampling layer by decomposing existing upsampling operations based on two steps:
1) methods of changing the spatial resolution of the feature map and
2) methods of feature transformation.
Our search space of operations for changing the spatial resolution includes: bilinear upsampling, bicubic upsampling~\cite{dong2015image}, nearest-neighbor interpolation, depth-to-space~\cite{shi2016real,ledig2017photo}, and stride $2$ transposed convolution~\cite{zeiler2011adaptive,long2015fully,dosovitskiy2015flownet}.
Our search space of operations for feature transformation includes: $2$D convolution, add every $N$ consecutive channels~\cite{Wojna2017TheDI,Wojna-2019}, separable convolution~\cite{Wojna2017TheDI,Wojna-2019,guo2018network}, depth-wise convolution~\cite{guo2019depthwise,gao2018channelnets,guo2018network}, and identity.
To relax the degree of freedom during the network architecture search, our search space allows the operations that contain learnable parameters to search for the kernel size, the dilation rate, and whether to include an activation at the end.
Our search space of operations for the activation function includes none, ReLU~\cite{nair2010rectified}, LeakyReLU~\cite{xu2015empirical}, SELU~\cite{klambauer2017self}, and PReLU~\cite{he2015delving}.
Figure~\ref{fig:upsample} presents our developed search space for the upsampling cell.
By decomposing several commonly used upsampling operators, our search space is more flexible and allows us to discover a novel upsampling cell for each task. 
Newly developed spatial upsampling operator (e.g., CARAFE~\cite{wang2019carafe}) can also be incorporated in our search space easily in the future.

\begin{figure}[!t]
  \begin{center}
  \mpage{0.3}{\includegraphics[width=\linewidth]{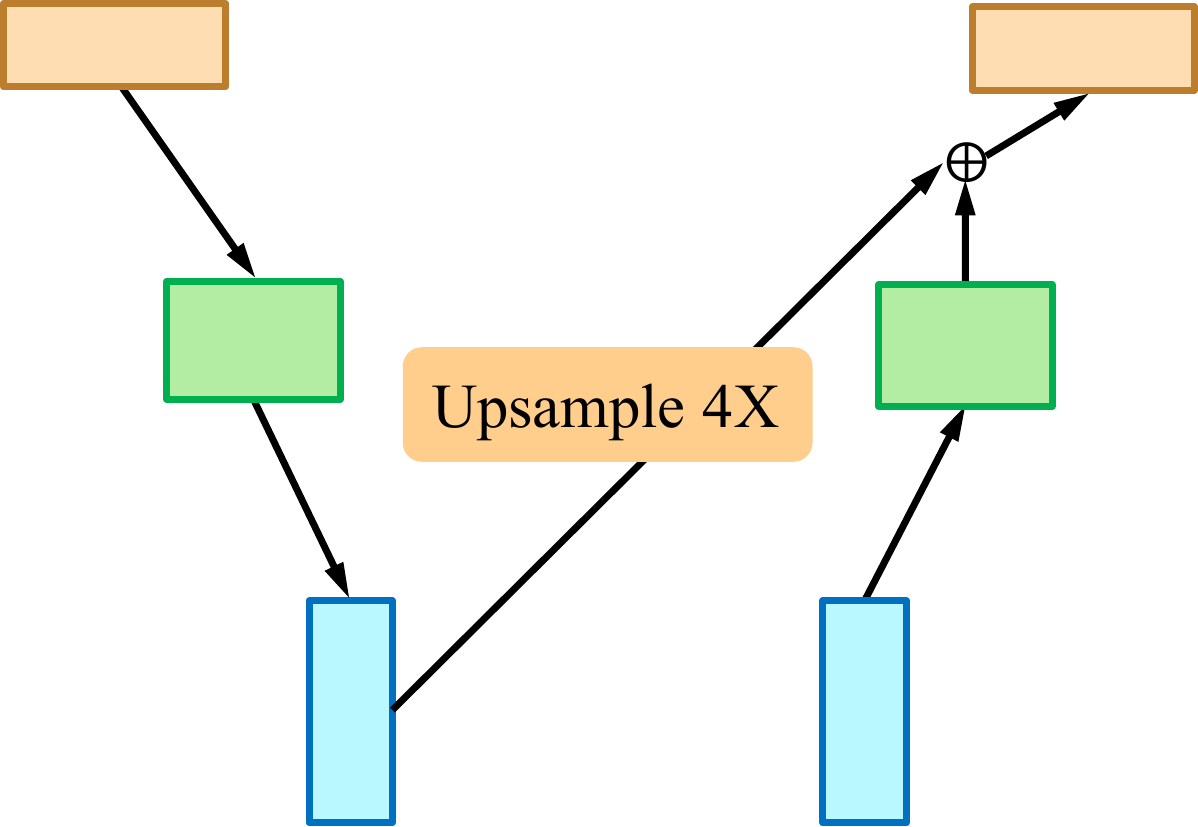}} \hfill
  \mpage{0.3}{\includegraphics[width=\linewidth]{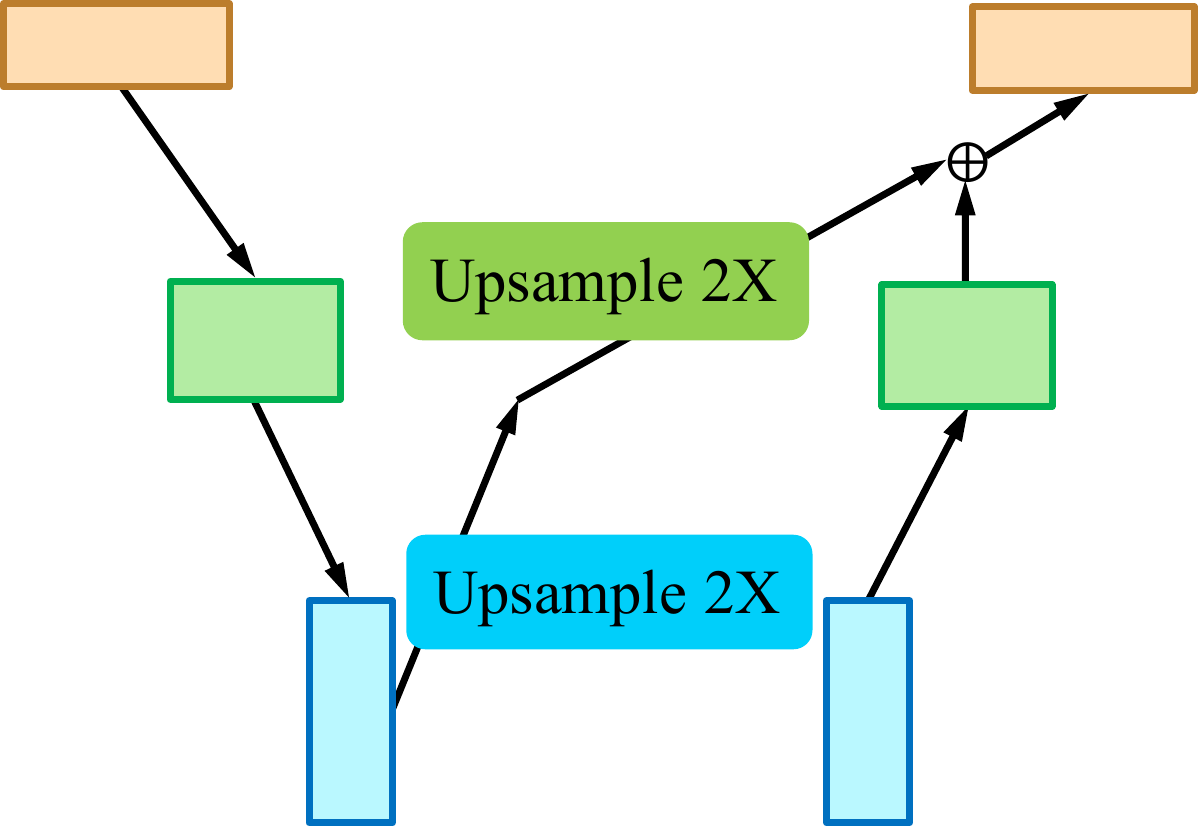}} \hfill
  \mpage{0.3}{\includegraphics[width=\linewidth]{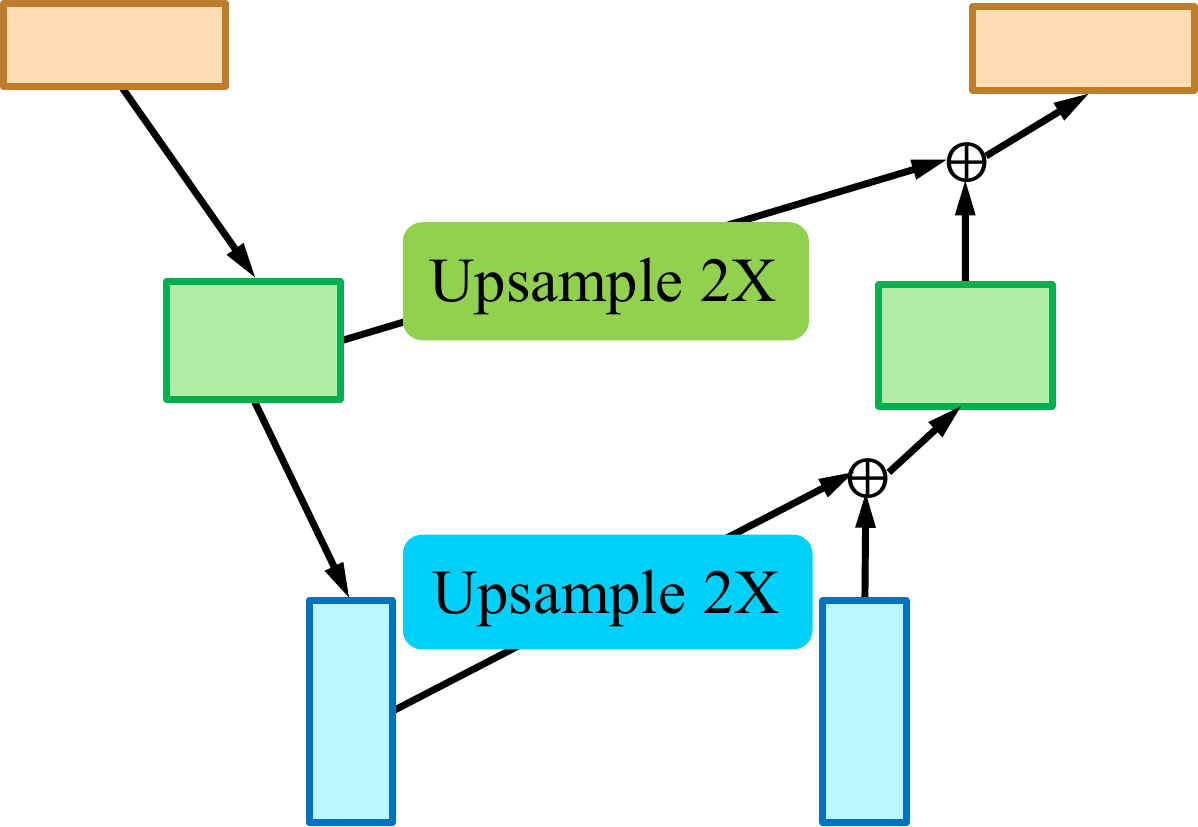}} \\
  \vspace{1.5mm}
  \mpage{0.3}{\small Direct upsampling} \hfill
  \mpage{0.3}{\small Progressive upsampling \\ (Ours)} \hfill
  \mpage{0.3}{\small Weight sharing \\ (Ours)} \\
  \vspace{\figcapmargin}
  \caption{
  \textbf{Decomposition and weight sharing of the upsampling operations.} 
  To achieve cross-scale residual connections, we decouple the upsampling operations into a series of $2\times$ upsampling operations, e.g., a $4\times$ upsampling operation (\emph{left}) can be realized by two consecutive $2\times$ upsampling operations (\emph{middle}).
  We adopt \emph{weight sharing} for the upsampling operation at the same feature level, i.e., the weights of the $2\times$ upsampling operations in the middle example are shared with those in the right example.
  After the cross-level feature upsampling, we add all the input feature maps at the same feature level. 
  The resulting feature map then becomes the input of the decoder at the next layer.
  We note that the same level feature connections always exist in our model.
  We do not visualize them in this figure due to presentation clarity.
  }
  \label{fig:cross-scale-upsample}
  \end{center}
  \vspace{2.0\figmargin}
\end{figure}

\subsection{Cross-scale residual connections}
For the cross-scale residual connections, we develop a search space that covers cross-level feature connections.
In contrast to U-Net~\cite{ronneberger2015u} which \emph{concatenates feature maps} at the same feature level, we adopt \emph{residual connections}~\cite{he2016deep}.
This allows us to merge feature maps extracted from different feature levels without the need to pre-define the number of input channels for each layer in the decoder because we design the number of input channels to always be the same. 

{\flushleft {\bf Sharing cross-level patterns.}}
The search space for the cross-scale residual connections can be extremely large. 
Assuming that the network depth is $d$ (i.e., there are $d$ feature levels in the encoder and $d$ feature levels in the decoder), the number of possible connection patterns is $2^{d^2}$.
To constrain the search space, we enforce the connection patterns to depend only on \emph{the difference of feature levels} (e.g., connecting all the feature maps to the feature maps one level higher).
For the $k$-th feature level in the decoder, it can receive feature maps from the $(k-d+1)$-th feature level up to the $(k+d-1)$-th feature level (i.e., there are in total $2^{2d-1}$ possible connection patterns).
With this constraint, we then search for a pattern of feature connections and \emph{share this pattern across different feature levels} in the decoder.
Figure~\ref{fig:connection-illustration} illustrates the main idea of the proposed cross-level residual connections.
The size of the search space can be significantly reduced from $2^{d^2}$ (without any constraints on the connection patterns) to $2^{2d-1}$.

{\flushleft {\bf Progressive upsampling.}}
For the cross-scale upsampling operation, as illustrated in Figure~\ref{fig:cross-scale-upsample}, we propose to \emph{decouple} the $4\times$ upsampling operation in the left into two consecutive $2\times$ upsampling operations in the middle, and share the weights with the $2\times$ upsampling operations in the right at the same feature level.
This allows us to define $2\times$ upsampling operations between each consecutive feature level only, and all other possible upsampling scales can be achieved by decoupling them into a series of $2\times$ upsampling operations.
The cross-scale downsampling connections can be achieved similarly.
\vspace{\secmargin}
\section{Experimental Results} \label{sec:results}
\vspace{0.5\secmargin}

In this section, we first describe the implementation details. 
We then present the quantitative and visual comparisons to existing methods as well as the ablation study.

\vspace{\secmargin}
\subsection{Implementation details}
Here, we provide the implementation details regarding the neural architecture search, model training, and testing.

{\flushleft {\bf Neural architecture search.}}
We implement our model using PyTorch.
The network architecture of our RNN controller is the same as \cite{Gong-2019}. 
To create the training data for searching for the network architecture, we randomly sample $100$ images from the DIV2K~\cite{agustsson2017ntire} training set.
To search for the optimal network structure for the task of interest, the neural architecture search process is composed of two alternating phases.
For the first phase, the RNN controller first samples a candidate network architecture with random initialization. 
We then optimize the sampled candidate model on the held-out training set (i.e., model training in our NAS-DIP framework).
For the second phase, we first compute the PSNR between the restored prediction and the corresponding ground truth as the reward and use reinforcement learning to update the RNN controller (i.e., NAS training in our NAS-DIP framework).
The training time required for each task varies. 
Specifically, finding the optimal network structure for the super-resolution task takes about $3$ days, denoising about $3$ days, and inpainting about $5$ days (using an NVIDIA V$100$ GPU with $12$GB memory).

{\flushleft {\bf Testing details.}}
When applying the searched model for testing, there is a hyperparameter (number of iterations) that one can set to obtain the final prediction results. 
Early method~\cite{DIP-CVPR-2018} uses the ground truth of the test image and the PSNR to select the number of iterations with the best performance.
However, this scheme may not be practical as the ground-truth image is not often available. 
To address this issue, we use the same training set for NAS training to find an optimal number of iterations that allows the model to reach the best performance. 
Specifically, we select model prediction at $4,500$ iterations for super-resolution, $3,500$ iterations for denoising, and $9,000$ iterations for inpainting.

\begin{table}[!t]
  \begin{center}
  \caption{
  \textbf{Comparison with existing methods on image restoration tasks.} 
  For (a), (b), and (c), we report the average PSNR results.
  For (d), we follow the evaluation protocol in~\cite{athar2018latent} and report the mean squared error (MSE).
  (a) Results of single image super-resolution on the Set5~\cite{Bevilacqua-2012} and Set14~\cite{Zeyde-2010} datasets with $2\times$, $4\times$, and $8\times$ scaling factors.
  (b) Results of image inpainting on the dataset provided by~\cite{Heide-CVPR-2015} (\emph{left}) and image denoising on the BM3D dataset~\cite{dabov2007video} (\emph{right}). 
  (c) Comparison with Deep Decoder~\cite{heckel2018deep} on the dataset provided by~\cite{heckel2018deep} on single image super-resolution and image inpainting tasks.
  (d) Comparison with Latent Convolutional Models~\cite{athar2018latent} on the Bedrooms dataset of LSUN~\cite{yu2015lsun}.
  For super-resolution in (c) and (d), we report the results with $4\times$ scaling factor.
  Marker $^*$ indicates that method uses the ground truth to get the best PSNR results.
  The results of all competing methods are from the respective papers.
  The \first{bold} and \second{underlined} numbers indicate the top two results, respectively.
  }
  \vspace{-2.0mm}
  \label{exp:restoration}
  \begin{minipage}[t]{0.69\textwidth}
    \scriptsize
    \subcaption{}
    \label{exp:dip-SR}
    \centering
    \resizebox{\linewidth}{!} 
    {
    \begin{tabular}{clcccccccc}
    \toprule
    \multirow{2}{*}{Type} & \multirow{2}{*}{Method} & \multicolumn{3}{c}{Set5~\cite{Bevilacqua-2012}} & & \multicolumn{3}{c}{Set14~\cite{Zeyde-2010}} \\
    & & $2\times$ & $4\times$ & $8\times$ & & $2\times$ & $4\times$ & $8\times$ \\
    \midrule
    \multirow{7}{*}{\rotatebox{90}{Learning based}} & LapSRN~\cite{lai2017deep} & 37.52 & 31.54 & 26.14 & & 33.08 & 28.19 & 24.44 \\
    & VDSR~\cite{kim2016accurate} & 37.53 & 31.35 & 25.93 & & 33.05 & 28.02 & 24.26 \\
    & EDSR~\cite{lim2017enhanced} & 38.11 & 32.46 & 26.96 & & 33.92 & 28.80 & 24.91 \\
    & RDN~\cite{zhang2018residual} & 38.24 & 32.47 & - & & 34.01 & 28.81 & - \\
    & RCAN~\cite{zhang2018image} & 38.27 & 32.63 & \second{27.31} & & \second{34.12} & 28.87 & \second{25.23} \\
    & SAN~\cite{dai2019second} & \second{38.31} & \second{32.64} & 27.22 & & 34.07 & \second{28.92} & 25.14 \\
    & EBRN~\cite{qiu2019embedded} & \first{38.35} & \first{32.79} & \first{27.45} & & \first{34.24} & \first{29.01} & \first{25.44} \\
    \midrule
    \multirow{9}{*}{\rotatebox{90}{Learning free}} & Bicubic~\cite{dong2015image} & 33.66 & 28.44 & 24.37 & & 30.24 & 26.05 & 23.09 \\
    & Glasner~\etal~\cite{irani2009super} &  - & 28.84 & - & & - & 26.46 & - \\
    & TV prior~\cite{beck2009fast} &  - & 28.85 & 24.87 & & - & 26.42 & 23.48 \\
    & RED~\cite{romano2017little} & - & 30.23 & 25.56 & & - & 27.36 & 23.89 \\
    & DeepRED~\cite{mataev2019deepred} & - & 30.72 & 26.04 & & - & 27.63 & 24.28 \\
    & SelfExSR~\cite{huang2015single} & \first{36.60} & 30.34 & 25.49 & & \first{32.24} & 27.41 & 23.92 \\
    & DIP$^*$~\cite{DIP-CVPR-2018} & 33.19 & 29.89 & 25.88 & & 29.80 & 27.00 & 24.15 \\
    & Ours & 35.32 & \second{30.81} & \second{26.41} & & 31.58 & \second{27.84} & \second{24.59} \\
    & Ours$^*$ & \second{35.90} & \first{31.09} & \first{27.03} & & \second{31.89} & \first{28.37} & \first{25.17} \\
    \bottomrule
    \end{tabular}
    }
  \end{minipage}
  \hfill
  \begin{minipage}[t]{0.27\textwidth}
    \scriptsize
    \subcaption{}
    \label{exp:dip-other}
    \centering
    \resizebox{\linewidth}{!} 
    {
    \begin{tabular}{l c c}
    \toprule
    Method & Inpainting & Denoising \\
    \midrule
    Papyan~\etal~\cite{papyan2017convolutional} & 31.19 & -\\
    DIP~\cite{DIP-CVPR-2018} & 33.48 & 30.43 \\ 
    SGLD~\cite{cheng2019bayesian} & \second{34.51} & \second{30.81} \\ 
    Ours & \first{34.72} & \first{31.42} \\
    \bottomrule
    \end{tabular}
    }
    \subcaption{}
    \label{exp:deepdecoder}
    \resizebox{\linewidth}{!} 
    {
    \begin{tabular}{l c c}
    \toprule
    Method & SR $4\times$ & Inpainting \\
    \midrule
    Deep Decoder~\cite{heckel2018deep} & 25.8 & 31.9 \\ 
    DIP~\cite{DIP-CVPR-2018} & \second{26.9} & \second{32.3} \\ 
    Ours & \first{27.4} & \first{33.1} \\
    \bottomrule
    \end{tabular}
    }
    \subcaption{}
    \label{exp:lcm}
    \resizebox{\linewidth}{!} 
    {
    \begin{tabular}{l c c}
    \toprule
    Method & SR $4\times$ $\downarrow$ & Inpainting $\downarrow$ \\
    \midrule
    PGAN~\cite{karras2017progressive} & 0.0183 & 0.0097 \\ 
    GLO~\cite{bojanowski2017optimizing} & 0.0069 & 0.0085 \\ 
    LCM~\cite{athar2018latent} & 0.0071 & 0.0065 \\ 
    DIP~\cite{DIP-CVPR-2018} & \second{0.0057} & \second{0.0063} \\ 
    Ours & \first{0.0054} & \first{0.0060} \\
    \bottomrule
    \end{tabular}
    }
  \end{minipage}
  \end{center}
  \vspace{1.0\tablemargin}
\end{table}

\subsection{Quantitative comparison}
We validate the effectiveness of our searched model architecture by evaluating its performance when used as a deep image prior to solve various inverse problems in image restoration. 
In each task, we compare with the state-of-the-art learning-free methods on benchmark datasets.

{\flushleft {\bf Single image super-resolution.}}
Following the evaluation protocols in DIP~\cite{DIP-CVPR-2018}, we adopt two standard benchmarks: the Set5~\cite{Bevilacqua-2012} and Set14~\cite{Zeyde-2010} datasets.
We compare our approach with existing learning-free methods~\cite{irani2009super,dong2015image,romano2017little,mataev2019deepred,huang2015single,DIP-CVPR-2018,beck2009fast} and learning-based approaches~\cite{lai2017deep,kim2016accurate,lim2017enhanced,zhang2018residual,zhang2018image,dai2019second,qiu2019embedded} on three different upsampling scales (i.e., $2\times$, $4\times$, and $8\times$ upsampling).

We use the evaluation code provided by DIP~\cite{DIP-CVPR-2018}.\footnote{\url{https://github.com/DmitryUlyanov/deep-image-prior}} 
Table~\ref{exp:dip-SR} presents the experimental results.
Results on all three upsampling scales show that our proposed algorithm performs favorably against the state-of-the-art \emph{learning-free} methods.
Our results show that for a larger upsampling scale (e.g., $8\times$ upsampling), our method achieves competitive or even better performance when compared with existing learning-based methods~\cite{lai2017deep,kim2016accurate} on both datasets.
This is interesting because our model has \emph{never seen any paired low-/high-resolution image pair}. 
The results also highlight the importance of our searched architecture, resulting in significant performance boost over existing methods that use hand-crafted priors (e.g., DIP~\cite{DIP-CVPR-2018}).

{\flushleft {\bf Image denoising and inpainting.}}
We adopt the BM3D dataset~\cite{dabov2007video} to evaluate the image denoising task.\footnote{\url{http://www.cs.tut.fi/~foi/GCF-BM3D/}} 
For fair comparison, we follow DIP~\cite{DIP-CVPR-2018} and average the output of our network using an exponential sliding window (moving average) to obtain the final result.
For evaluating the performance of the inpainting task, we follow DIP~\cite{DIP-CVPR-2018} and use the dataset provided by Heide~\etal~\cite{Heide-CVPR-2015}.\footnote{\url{https://dmitryulyanov.github.io/deep_image_prior}}
Here, we compare our results with DIP~\cite{DIP-CVPR-2018}, Papyan~\etal~\cite{papyan2017convolutional}, and SGLD~\cite{cheng2019bayesian} using the $50\%$ missing pixels setting.
Table~\ref{exp:dip-other} reports the experimental results. 
Similarly, our method compares favorably against all competing approaches on both tasks.

\begin{figure}[!t]
  \begin{center}
  \mpage{0.1795}{\includegraphics[width=\linewidth]{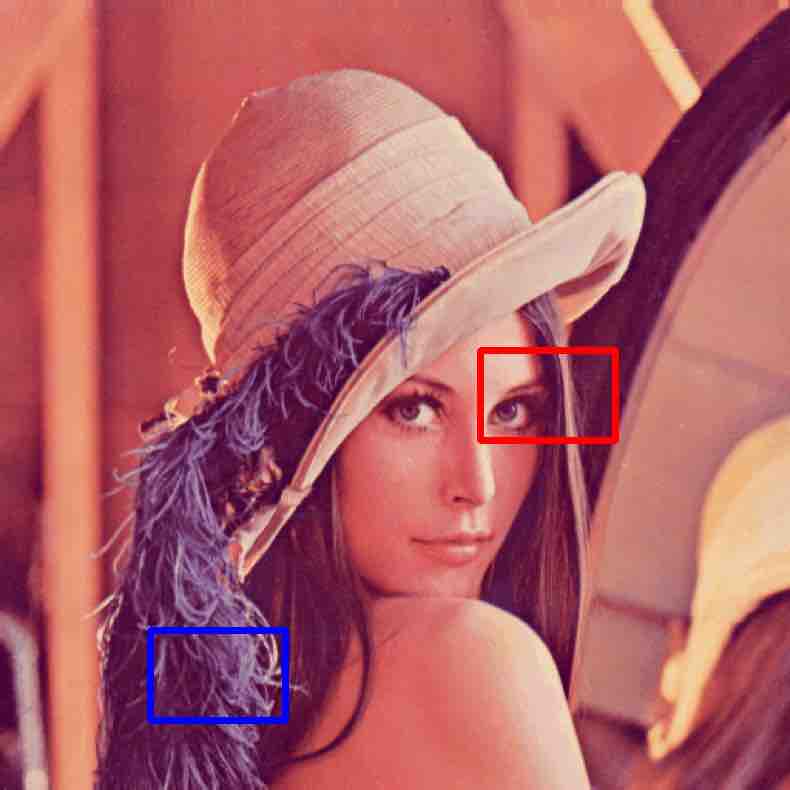}} \hfill
  \mpage{0.1795}{\includegraphics[width=\linewidth]{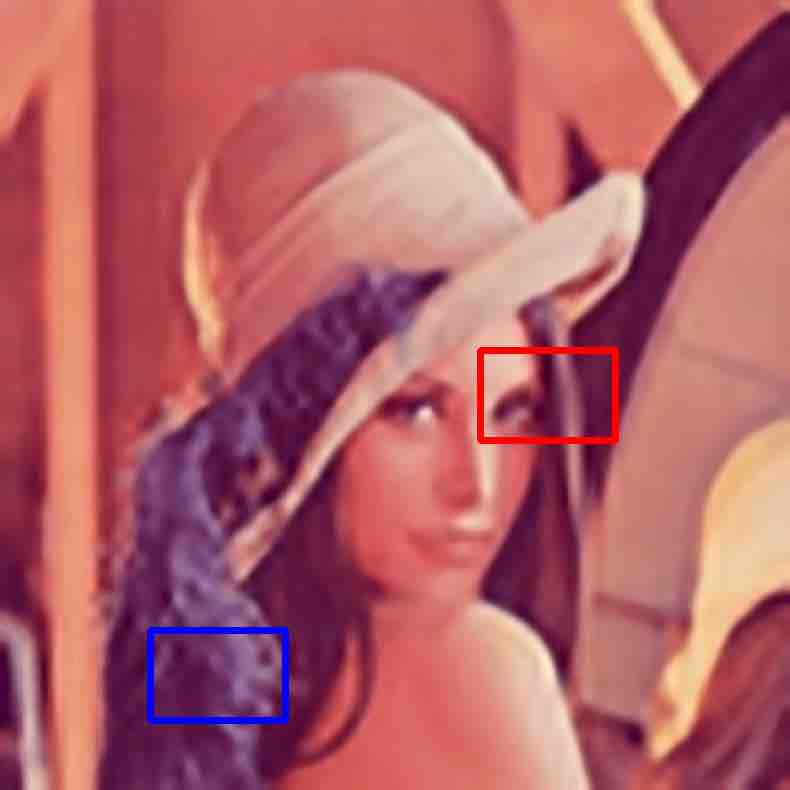}} \hfill
  \mpage{0.1795}{\includegraphics[width=\linewidth]{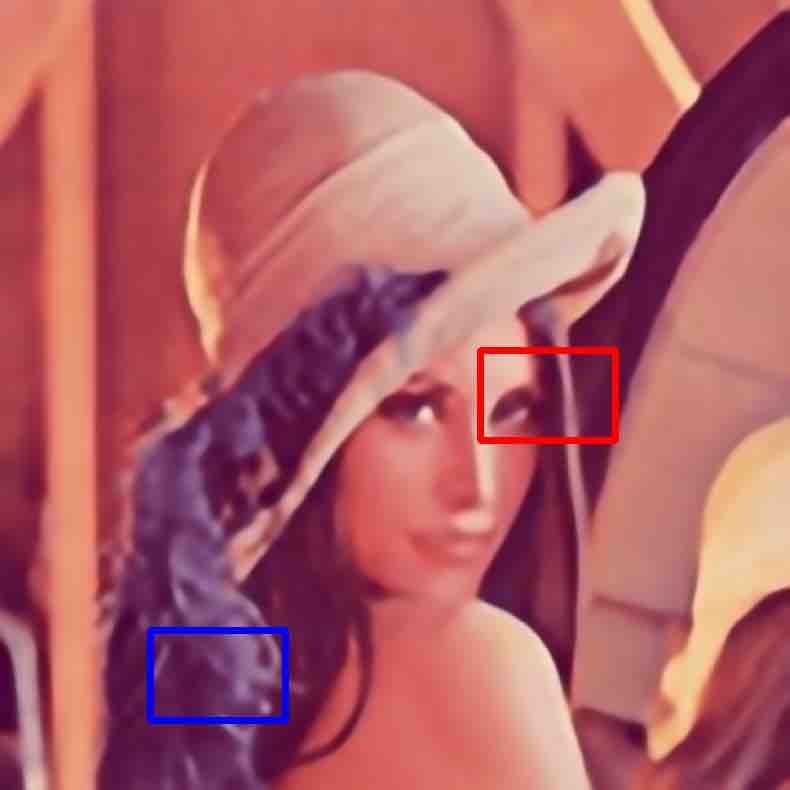}} \hfill
  \mpage{0.1795}{\includegraphics[width=\linewidth]{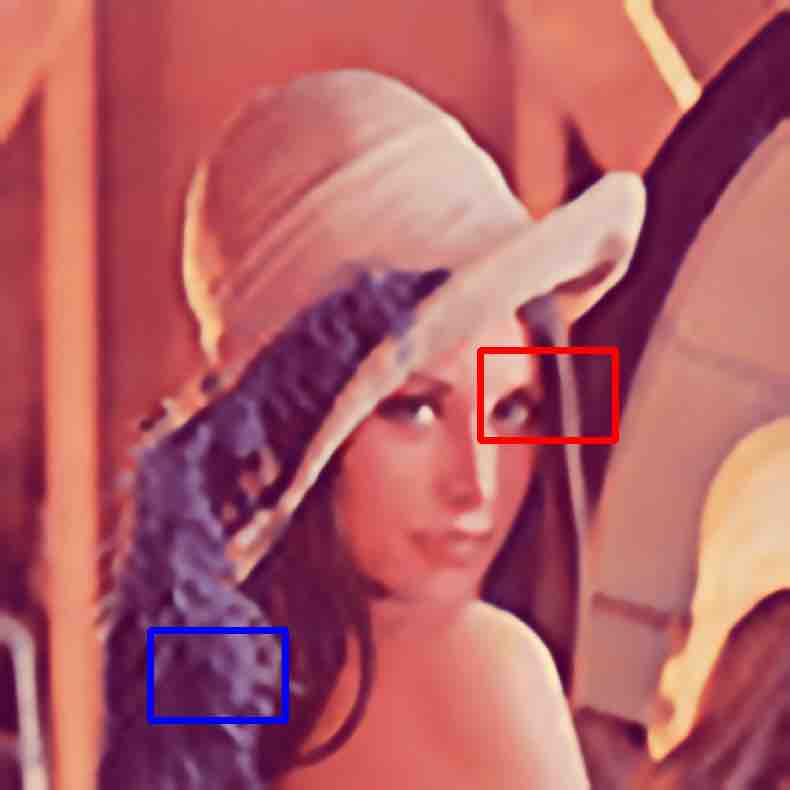}} \hfill
  \mpage{0.1795}{\includegraphics[width=\linewidth]{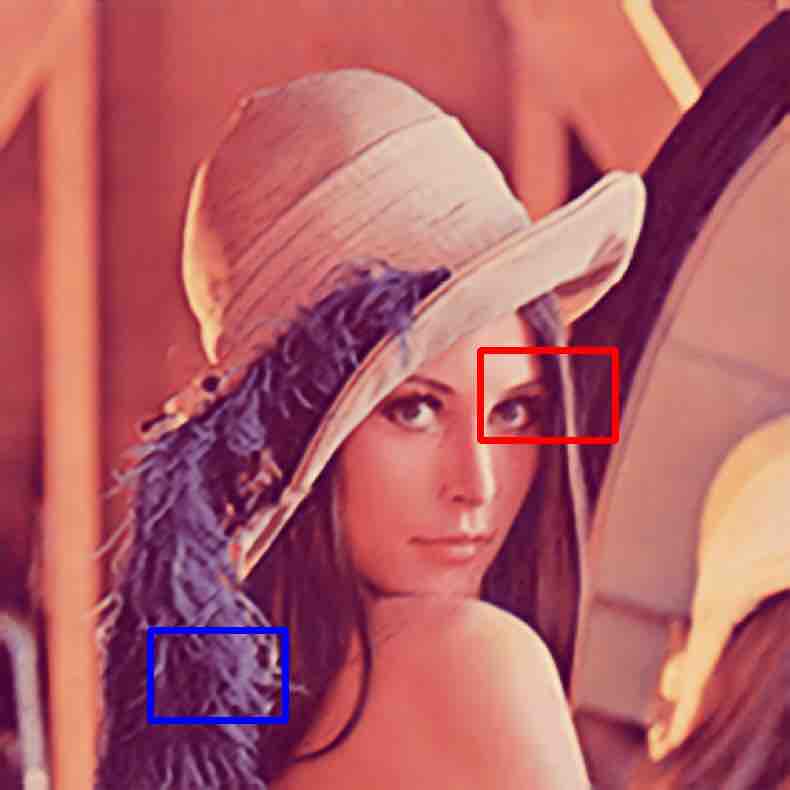}} \\
  \mpage{0.1795}{\includegraphics[width=0.47\linewidth]{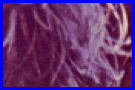} \hfill \includegraphics[width=0.47\linewidth]{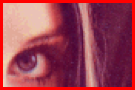}} \hfill
  \mpage{0.1795}{\includegraphics[width=0.47\linewidth]{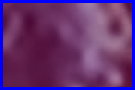} \hfill \includegraphics[width=0.47\linewidth]{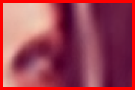}} \hfill
  \mpage{0.1795}{\includegraphics[width=0.47\linewidth]{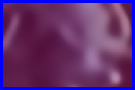} \hfill \includegraphics[width=0.47\linewidth]{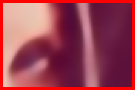}} \hfill
  \mpage{0.1795}{\includegraphics[width=0.47\linewidth]{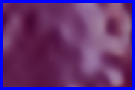} \hfill \includegraphics[width=0.47\linewidth]{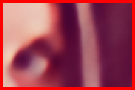}} \hfill
  \mpage{0.1795}{\includegraphics[width=0.47\linewidth]{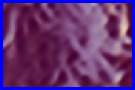} \hfill \includegraphics[width=0.47\linewidth]{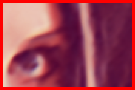}} \\
  \vspace{1.0mm}
  \mpage{0.1795}{Ground truth} \hfill
  \mpage{0.1795}{Bicubic} \hfill
  \mpage{0.1795}{DIP~\cite{DIP-CVPR-2018}} \hfill
  \mpage{0.1795}{LapSRN~\cite{lai2017deep}} \hfill
  \mpage{0.1795}{Ours} \\
  \vspace{\figcapmargin}
  \caption{
  \textbf{Qualitative results of single image super-resolution.} 
  We present visual comparisons with learning-free methods (i.e., bicubic and DIP~\cite{DIP-CVPR-2018}) and a learning-based approach (i.e., LapSRN~\cite{lai2017deep}) with $8\times$ scaling factor.
  }
  \label{fig:SR}
  \end{center}
  \vspace{1.0\figmargin}
\end{figure}

\begin{figure}[!t]
  \begin{center}
  \mpage{0.23}{\includegraphics[width=\linewidth]{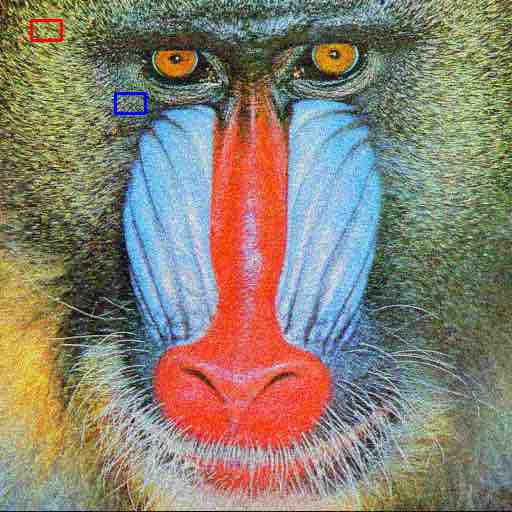}} \hfill
  \mpage{0.23}{\includegraphics[width=\linewidth]{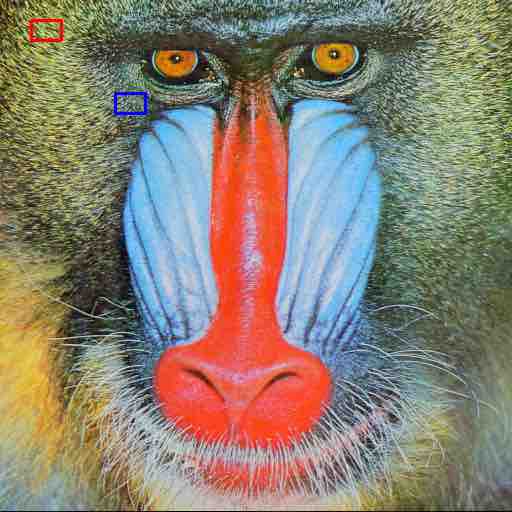}} \hfill
  \mpage{0.23}{\includegraphics[width=\linewidth]{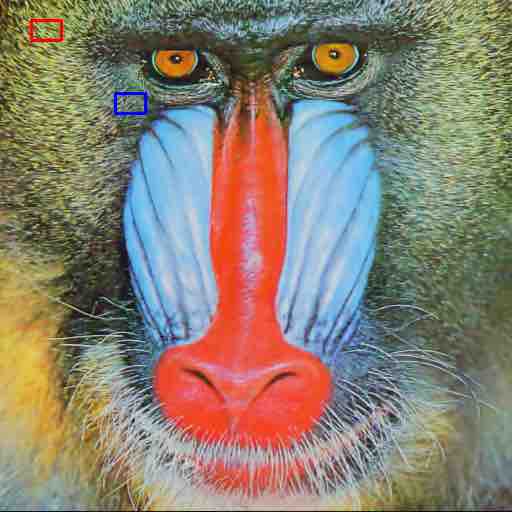}} \hfill
  \mpage{0.23}{\includegraphics[width=\linewidth]{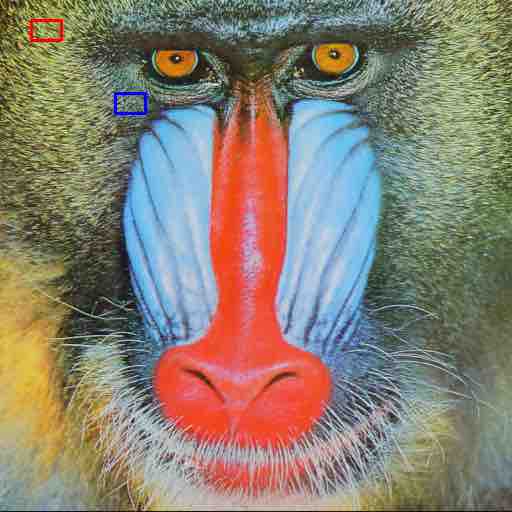}} \\
  \mpage{0.23}{\includegraphics[width=0.47\linewidth]{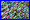} \hfill \includegraphics[width=0.47\linewidth]{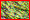}} \hfill
  \mpage{0.23}{\includegraphics[width=0.47\linewidth]{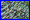} \hfill \includegraphics[width=0.47\linewidth]{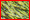}} \hfill
  \mpage{0.23}{\includegraphics[width=0.47\linewidth]{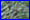} \hfill \includegraphics[width=0.47\linewidth]{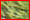}} \hfill
  \mpage{0.23}{\includegraphics[width=0.47\linewidth]{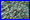} \hfill \includegraphics[width=0.47\linewidth]{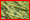}} \\
  \vspace{1.0mm}
  \mpage{0.23}{Noisy image} \hfill
  \mpage{0.23}{Ground truth} \hfill
  \mpage{0.23}{DIP~\cite{DIP-CVPR-2018}} \hfill
  \mpage{0.23}{Ours} \\
  \vspace{\figcapmargin}
  \caption{
  \textbf{Qualitative results of image denoising.} 
  We present the visual comparisons with DIP~\cite{DIP-CVPR-2018} on the BM3D dataset~\cite{dabov2007video}.
  }
  \label{fig:denoising}
  \end{center}
  \vspace{2.0\figmargin}
\end{figure}

{\flushleft {\bf Comparisons to recent CNN designs.}}
There have been several recent methods that design the CNN architecture for improved performance on image restoration tasks. 
We first follow the same experimental setting as DeepDecoder~\cite{heckel2018deep} and evaluate our method on the $4\times$ super-resolution and inpainting tasks.\footnote{\url{https://github.com/reinhardh/supplement_deep_decoder}}
Table~\ref{exp:deepdecoder} reports the experimental results.
Next, we follow the evaluation protocol in Latent Convolutional Models~\cite{athar2018latent} and report our results on the $4\times$ super-resolution and inpainting tasks in Table~\ref{exp:lcm}.\footnote{\url{https://github.com/srxdev0619/Latent_Convolutional_Models}}

From extensive quantitative evaluations, we show that our model with both the searched upsampling cells and the cross-scale residual connections can serve as a stronger structured image prior to existing manual CNN architecture designs.

\subsection{Visual comparison}
Here, we show sample qualitative results of several image restoration tasks and compare them with the state-of-the-art approaches.
We refer the reader to review the full resolution results to better perceive visual quality improvement.

Figure~\ref{fig:SR} and Figure~\ref{fig:denoising} present the visual results of single image super-resolution and image denoising, respectively.
Generally, using our model as an image prior results in clearly visible improvement in terms of the visual quality. 
This improvement highlights the strength of \emph{learning} a stronger structured image prior through neural architecture search.

\begin{figure}[!t]
  \begin{center}
  \mpage{0.23}{\includegraphics[width=\linewidth]{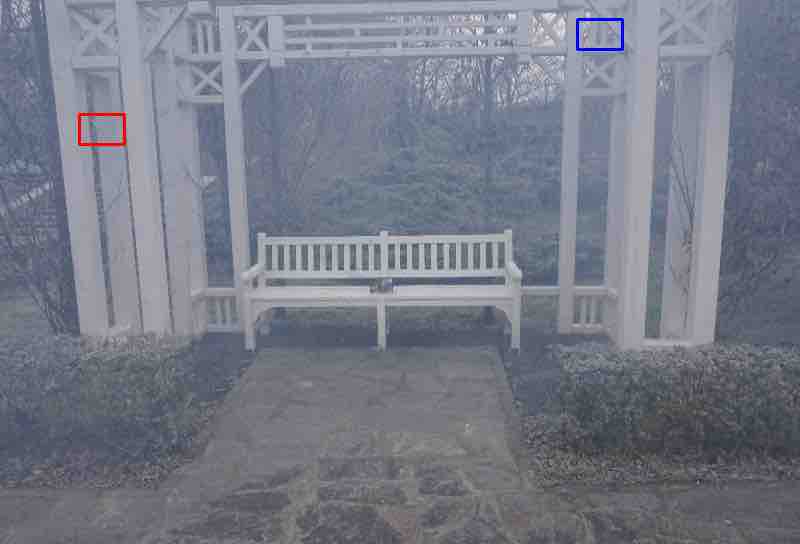}} \hfill
  \mpage{0.23}{\includegraphics[width=\linewidth]{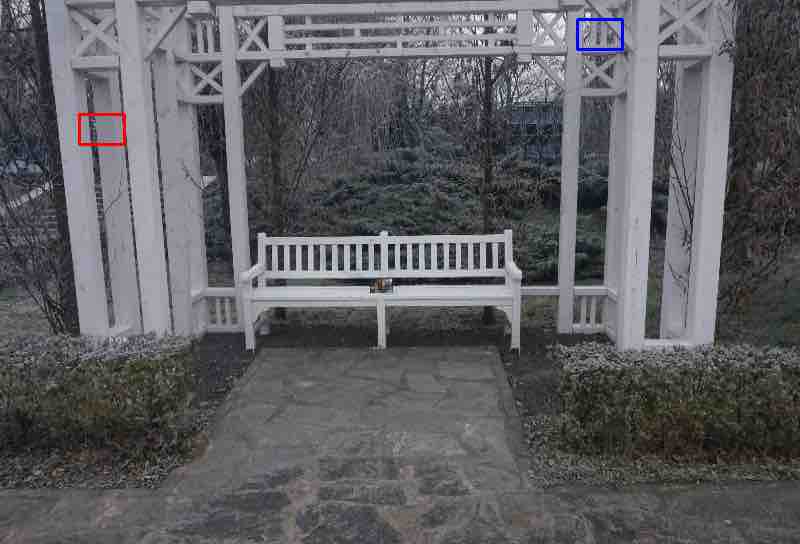}} \hfill
  \mpage{0.23}{\includegraphics[width=\linewidth]{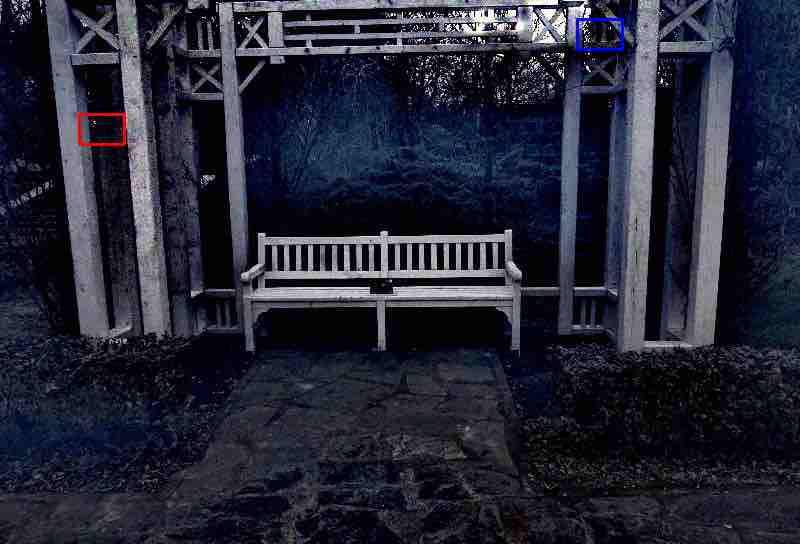}} \hfill
  \mpage{0.23}{\includegraphics[width=\linewidth]{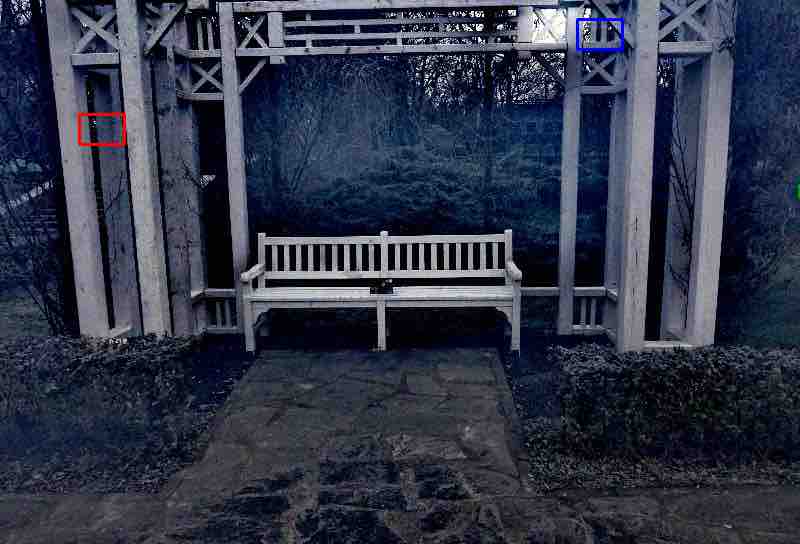}} \\
  \mpage{0.23}{\includegraphics[width=0.47\linewidth]{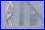} \hfill \includegraphics[width=0.47\linewidth]{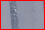}} \hfill
  \mpage{0.23}{\includegraphics[width=0.47\linewidth]{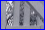} \hfill \includegraphics[width=0.47\linewidth]{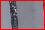}} \hfill
  \mpage{0.23}{\includegraphics[width=0.47\linewidth]{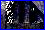} \hfill \includegraphics[width=0.47\linewidth]{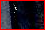}} \hfill
  \mpage{0.23}{\includegraphics[width=0.47\linewidth]{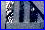} \hfill \includegraphics[width=0.47\linewidth]{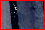}} \\
  \vspace{1.0mm}
  \mpage{0.23}{\small Hazy image} \hfill
  \mpage{0.23}{\small Ground truth} \hfill
  \mpage{0.23}{\small DoubleDIP (U-Net)} \hfill
  \mpage{0.23}{\small DoubleDIP (Ours)} \\
  \vspace{\figcapmargin}
  \caption{
  \textbf{Qualitative results of image dehazing.} 
  We present the visual comparisons with DoubleDIP~\cite{gandelsman2018double} on the O-HAZE dataset~\cite{ancuti2018haze}.
  }
  \label{fig:dehazing}
  \end{center}
  \vspace{2.5\figmargin}
\end{figure}

In addition to standard image restoration tasks, we also experiment with \emph{model transferability} to two different tasks. 
We use the dehazing application in DoubleDIP~\cite{gandelsman2018double} and the matrix factorization task in CompMirror~\cite{aittala2019computational} for demonstration.

For dehazing, we follow the official implementation by DoubleDIP~\cite{gandelsman2018double} for generating the dehazed results.\footnote{\url{https://github.com/yossigandelsman/DoubleDIP}}
To generate our results, we swap the standard U-Net model in DoubleDIP~\cite{gandelsman2018double} with our searched model searched in the \emph{denoising} task.
Figure~\ref{fig:dehazing} shows an example of visual comparison with DoubleDIP~\cite{gandelsman2018double} on the O-HAZE dataset~\cite{ancuti2018haze}.\footnote{We originally plan to conduct a quantitative evaluation using the O-HAZE dataset~\cite{ancuti2018haze}. Unfortunately, using the provided source code and email correspondences with the authors, we were still unable to reproduce the results of DoubleDIP on this dataset. We thus did not report quantitative results on dehazing in this work.}
Our results show that using our model produces dehazed images with better visual quality.

\begin{figure}[!t]
  \begin{center}
  \mpage{0.316}{\includegraphics[height=0.32\linewidth]{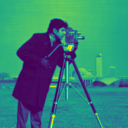} \includegraphics[height=0.32\linewidth]{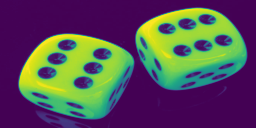}} \hfill
  \mpage{0.316}{\includegraphics[height=0.32\linewidth]{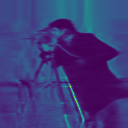}  \includegraphics[height=0.32\linewidth]{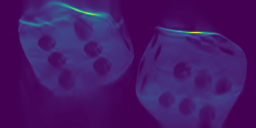}} \hfill
  \mpage{0.316}{\includegraphics[height=0.32\linewidth]{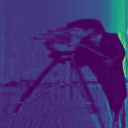} \includegraphics[height=0.32\linewidth]{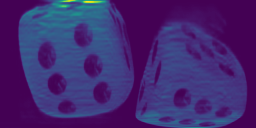}} \\
  \vspace{1.0mm}
  \mpage{0.316}{Input} \hfill
  \mpage{0.316}{CompMirror~\cite{aittala2019computational}} \hfill
  \mpage{0.316}{CompMirror (Ours)} \\
  \vspace{\figcapmargin}
  \caption{
  \textbf{Qualitative results of the matrix factorization.} 
  We present the visual comparisons with CompMirror~\cite{aittala2019computational} on the matrix factorization task.
  }
  \label{fig:compmirror}
  \end{center}
  \vspace{2.0\figmargin}
\end{figure}

For matrix factorization, we use the official implementation by CompMirror~\cite{aittala2019computational} to generate the factorized results.\footnote{\url{https://github.com/prafull7/compmirrors}}
To generate our results, we replace the upsampling layer in the CompMirror~\cite{aittala2019computational} model with our searched upsampling layer searched in the \emph{denoising} task.
Figure~\ref{fig:compmirror} presents an example of visual comparison with CompMirror~\cite{aittala2019computational}.
Our results show that using our model produces smoother factorized images with fewer visual artifacts.

\begin{table}[!t]
  \begin{center}
  \scriptsize
  \caption{
  \textbf{Ablation study.}
  We report the average PSNR results with comparisons to our variant methods.
  For single image super-resolution, we report the results on the Set 14 dataset~\cite{Zeyde-2010} with $4\times$ and $8\times$ scaling factors.
  For image inpainting, we report the results using the dataset provided by \cite{Heide-CVPR-2015}.
  For image denoising, we use the BM3D dataset~\cite{dabov2007video} for evaluation.
  We denote ``S-U'' for search upsampling and ``S-C'' for search connection.
  The numbers in the parenthesis denote the performance gain over DIP~\cite{DIP-CVPR-2018}.
  The \first{bold} and \second{underlined} numbers indicate the top two results, respectively.
  }
  \vspace{1.0mm}
  \label{exp:ablation}
  \resizebox{\linewidth}{!} 
  {
  \begin{tabular}{lcc|cccc}
  \toprule
  Method & Search upsampling & Search connection & SR $4\times$ & SR $8\times$ & Inpainting & Denoising \\
  \midrule
  DIP~\cite{DIP-CVPR-2018} & - & - & 27.00 & 24.15 & 33.48 & 30.43 \\
  Ours w/o S-C & \checkmark & - & \second{27.54} (+0.54) & \second{24.44} (+0.29) & 34.01 (+0.53) & 31.08 (+0.65) \\
  Ours w/o S-U & - & \checkmark & 27.32 (+0.32) & 24.29 (+0.14) & \second{34.59} (+1.11) & \second{31.16} (+0.73) \\
  Ours & \checkmark & \checkmark & \first{27.84} (+0.84) & \first{24.59} (+0.44) & \first{34.72} (+1.24) & \first{31.42} (+0.99) \\
  \bottomrule
  \end{tabular}
  }
  \end{center}
  \vspace{2.0\tablemargin}
\end{table}

\vspace{1.5\secmargin}
\subsection{Ablation study}
\vspace{\secmargin}

We conduct an ablation study to isolate the contributions from individual components. 
Specifically, we aim to understand how much performance improvement can be attributed to each of our two technical contributions.
As our method builds upon the U-Net architecture of DIP~\cite{DIP-CVPR-2018}, we use their results as the baseline. 

We report the results of our variant methods in Table~\ref{exp:ablation}.
Our results demonstrate that searching for an upsampling cell and a pattern of cross-level residual connections consistently helps improve the performance over DIP~\cite{DIP-CVPR-2018} across multiple tasks. 
We also observe that the upsampling cell is particularly important for the single image super-resolution task.
On the other hand, introducing the searched cross-scale residual connections offers a larger performance boost over upsampling cell for both inpainting and denoising tasks.
The model with both components shows the best performance, highlighting the complementary nature of these two components.

\vspace{\secmargin}
\subsection{Image-to-image translation}

\begin{figure}[!t]
  \begin{center}
  \mpage{0.48}{Winter $\rightarrow$ Summer} \hfill
  \mpage{0.48}{Summer $\rightarrow$ Winter} \\
  \vspace{1.0mm}
  \mpage{0.145}{\includegraphics[width=\linewidth]{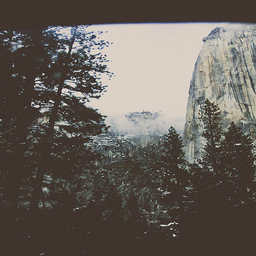}} \hfill
  \mpage{0.145}{\includegraphics[width=\linewidth]{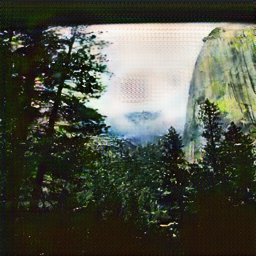}} \hfill
  \mpage{0.145}{\includegraphics[width=\linewidth]{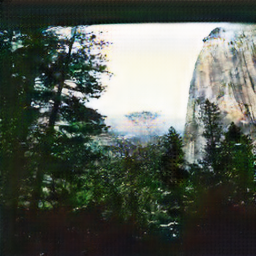}} \hfill
  \mpage{0.145}{\includegraphics[width=\linewidth]{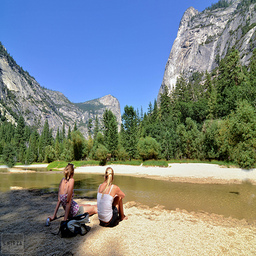}} \hfill
  \mpage{0.145}{\includegraphics[width=\linewidth]{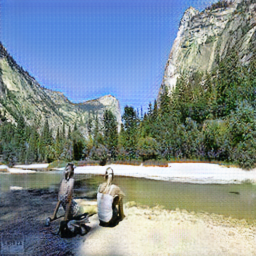}} \hfill
  \mpage{0.145}{\includegraphics[width=\linewidth]{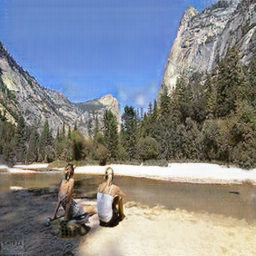}} \\
  \vspace{1.0mm}
  \mpage{0.145}{\scriptsize Input} \hfill
  \mpage{0.145}{\scriptsize CycleGAN (U-Net of \cite{zhu2017unpaired})} \hfill
  \mpage{0.145}{\scriptsize CycleGAN (Ours)} \hfill 
  \mpage{0.145}{\scriptsize Input} \hfill 
  \mpage{0.145}{\scriptsize CycleGAN (U-Net of \cite{zhu2017unpaired})} \hfill
  \mpage{0.145}{\scriptsize CycleGAN (Ours)} \\
  \vspace{\figcapmargin}
  \caption{
  \textbf{Qualitative results of the unpaired image-to-image translation task.} 
  We present the visual comparisons with CycleGAN~\cite{zhu2017unpaired} on the Winter $\rightarrow$ Summer (\emph{left}) and the Summer $\rightarrow$ Winter (\emph{right}) translation tasks.
  }
  \label{fig:cycleGAN}
  \end{center}
  \vspace{2.0\figmargin}
\end{figure}

We also explore transferring the searched model (from the denoising task) to a different problem. 
Specifically, we aim to test if our searched model generalizes well to image-to-image translation tasks.
We take the PyTorch implementation of CycleGAN~\cite{zhu2017unpaired} provided by the author and train the Summer $\leftrightarrow$ Winter translation.\footnote{\url{https://github.com/junyanz/pytorch-CycleGAN-and-pix2pix}} 
We compare our results with the standard U-Net based model of CycleGAN~\cite{zhu2017unpaired}. 
Figure~\ref{fig:cycleGAN} shows one sample result for each of the translation directions. 
To quantify the performance, we also compute the FID score~\cite{heusel2017gans} and perform a user study.
We report the results in Table~\ref{exp:summer2winter}. 
Both objective (FID score) and subjective (user study) results indicate that our searched model improves the performance over the base CycleGAN~\cite{zhu2017unpaired} model.

\begin{table}[!t]
  \begin{center}
  \caption{
  \textbf{Quantitative results of unpaired image-to-image translation on the Summer $\leftrightarrow$ Winter dataset.} 
  (\emph{Left}) The FID scores. 
  (\emph{Right}) The user study results.
  }
  \vspace{1.0mm}
  \label{exp:summer2winter}
  \begin{minipage}[t]{0.495\textwidth}
    \scriptsize
    \centering
    \resizebox{\linewidth}{!}
    {
    \begin{tabular}{lcc}
    \toprule
    \multirow{2}{*}{Method} & \multicolumn{2}{c}{FID $\downarrow$} \\ 
    & Summer $\rightarrow$ Winter & Winter $\rightarrow$ Summer \\ 
    \midrule
    CycleGAN (U-Net of \cite{zhu2017unpaired}) & 78.62 & 73.91  \\
    CycleGAN (U-Net of \cite{DIP-CVPR-2018}) & 79.74 & 74.83 \\
    CycleGAN (Ours) & 76.22 & 71.98  \\ 
    \bottomrule
    \end{tabular}
    }
  \end{minipage}
  \hfill
  \begin{minipage}[t]{0.495\textwidth}
    \scriptsize
    \centering
    \resizebox{\linewidth}{!}
    {
    \begin{tabular}{lcc}
    \toprule
    \multirow{2}{*}{Method} & \multicolumn{2}{c}{User study} \\ 
    & Summer $\rightarrow$ Winter & Winter $\rightarrow$ Summer \\ 
    \midrule
    CycleGAN (U-Net of \cite{zhu2017unpaired}) & 20.7\% & 27.82\% \\
    CycleGAN (Ours) & 79.3\% & 72.18\% \\ 
    \bottomrule
    \end{tabular}
    }
  \end{minipage}
  \end{center}
  \vspace{2.0\tablemargin}
\end{table}
\vspace{1.5\secmargin}
\section{Conclusions}
\vspace{1.0\secmargin}

In this paper, we propose to use neural architecture search techniques to discover stronger structured image priors captured by the CNN architecture. 
The core technical contributions of our work lie in (1) the search space design for the upsampling layer commonly used in a decoder and (2) the cross-level feature connections between the encoder and the decoder. 
We build our network design upon the standard U-Net architecture and search for an optimal upsampling cell and a pattern of cross-level feature connections for each task of interest.  
We validate the effectiveness of our model on four image restoration tasks, one matrix factorization application, and an unpaired image-to-image translation problem.
Through extensive experimental evaluations, our results show consistent performance improvement over conventional network designs.

\clearpage
%
%
\bibliographystyle{splncs04}
\bibliography{reference.bib}
\end{document}